\documentclass[10pt,letterpaper]{flair}

\usepackage{natbib}
\newcommand{\KL}{D_{\mathrm{KL}}}

\title{recAMPLIFY: Large-Scale Pretraining of Recursive Protein Language Models}
\runningtitle{recAMPLIFY: Large-Scale Pretraining of Recursive Protein Language Models}

\author[1,2]{Saurav Jha}
\author[3,4]{Qianli Liao}
\author[5]{Robert Vernon}
\author[1,2,6]{Sarath Chandar}
\author[1,7]{Quentin~Fournier}
\affil[1]{Mila -- Quebec AI Institute}
\affil[2]{Polytechnique Montréal}
\affil[3]{Center for Brains, Minds, and Machines}
\affil[4]{MIT}
\affil[5]{Amgen}
\affil[6]{Canada CIFAR AI Chair}
\affil[7]{University of Montréal}
\correspondingauthor{Saurav Jha (\email{saurav.jha@polymtl.ca}) and
    Quentin Fournier (\email{quentin.fournier@mila.quebec})}

\newcommand{\seq}{S}                             
\newcommand{\seqlen}{T}                          
\newcommand{\resi}[1][i]{a_{#1}}                 
\newcommand{\dmodel}{D}                          
\newcommand{\emb}{\mathbf{x}}                    
\newcommand{\prelude}{\mathrm{p}}                
\newcommand{\coda}{\mathrm{c}}                   
\newcommand{\fcore}{f_r}                         
\newcommand{\preludefn}{f_{\prelude}}            
\newcommand{\codafn}{f_{\coda}}                  
\newcommand{\best}[1]{\mathbf{#1}}
\newcommand{\zlat}{\mathbf{z}}                   
\newcommand{\zinit}{\mathbf{z}_\text{init}}      
\newcommand{\zstep}[1][k]{\mathbf{z}_{#1}}       
\newcommand{\ncyc}{L}                            
\newcommand{\kcyc}{k}                            
\newcommand{\effdepth}{\mathcal{D}_{\mathrm{eff}}}
\newcommand{\pred}[1][k]{\hat{\mathbf{y}}_{#1}}  
\newcommand{\halthead}{\texttt{HaltHead}}        
\newcommand{\lam}[1][k]{\lambda_{#1}}            
\newcommand{\phalt}[1][k]{p_{#1}}                
\newcommand{\Loss}{\mathcal{L}}
\newcommand{\Lmlm}{\Loss_{\mathrm{MLM}}}         
\newcommand{\Embed}{\mathrm{Embed}}
\newcommand{\stopgrad}{\mathrm{sg}}              
\newcommand{\En}{\mathbb{E}[k]}                  

\usepackage{wrapfig}
\usepackage{afterpage}
\usepackage{subcaption}
\usepackage{makecell}
\usepackage{adjustbox}
\usepackage{listings}

\lstdefinestyle{pseudocode}{
    language=Python,
    basicstyle=\ttfamily\scriptsize,
    keywordstyle=\bfseries\color{blue!45!black},
    commentstyle=\itshape\color{black!55},
    stringstyle=\color{purple!60!black},
    morekeywords={with,no_grad,stop_grad,sigmoid,Norm,Prelude,Embed,Coda,LM_head},
    mathescape=true,
    columns=fullflexible,
    keepspaces=true,
    showstringspaces=false,
    breaklines=true,
    breakatwhitespace=true,
    tabsize=2,
    frame=single,
    rulecolor=\color{black!20},
    backgroundcolor=\color{black!2},
    xleftmargin=0.6em,
    xrightmargin=0.3em,
    framexleftmargin=0.4em,
    aboveskip=0.5em,
    belowskip=0.2em,
    captionpos=b
}

\titleformat{\paragraph}[runin]
{\color{FlairPurple}\bfseries}{}{0em}{#1}[\quad]

\begin{abstract}
  Protein language models (pLMs) are powerful tools for protein annotation and mutation-effect prediction, but progress has largely depended on scaling model size and data, making them expensive and prone to memorization. Test-time scaling offers a complementary path, but methods developed for autoregressive LLMs do not transfer directly: masked encoders lack a generation loop, and amino acids do not verbalize in a reasoning language. Instead, we adapt recursive pretraining to masked pLMs, where a weight-tied Transformer core refines residue-aligned hidden states over an adjustable number of cycles. We instantiate this recipe as recAMPLIFY, a pLM with $155$M unique parameters trained on the same $2$T tokens as AMPLIFY-$350$M. Despite $56\%$ fewer parameters, it improves CASP contact recovery and three out of four supervised tasks, notably raising fold accuracy by $22.5\%$. Because optimal depth for recAMPLIFY varies across tasks and assays, we then introduce \emph{Gated Mixture Recurrence} (GMR), a framework that decodes each protein from a probability-weighted mixture of hidden states visited during recursion. GMR employs a learned residual gate at scalar, low-rank, or per-dimension resolution to control how much of each cycle's proposed update is accepted into the state. This unifies halting and gating into a single adaptive computation objective while recovering PonderNet-style halting as a special case without gates.  Layered on top of recAMPLIFY, GMR further reduces expected layer depth by up to $37\%$, although the accuracy trade-off varies by task. Mechanistic analyses show that residual gating controls recurrence stability beyond the trained depth and that halt depth tracks biological properties.

\includegraphics[height=1em]{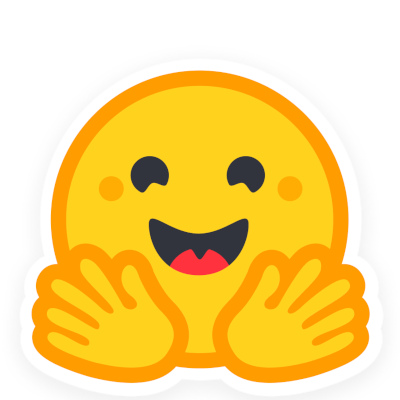}
\textbf{Model Weights}:
\href{https://huggingface.co/chandar-lab/recamplify}{\textbf{huggingface.co/chandar-lab/recamplify}}

\includegraphics[height=1em]{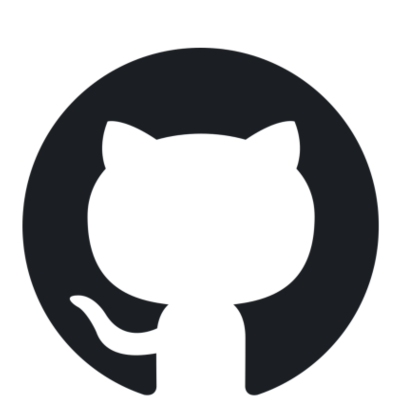}
\textbf{Code Repository}:
\href{https://github.com/chandar-lab/AMPLIFY/tree/recamplify}{\textbf{github.com/chandar-lab/AMPLIFY/tree/recamplify}}
\end{abstract}

\begin{document}
\maketitle

\section{Introduction}\label{sec:introduction}

Proteins are sequences of amino acids shaped by evolutionary pressure to code diverse biological functions, including the ability to form specific folded structures. Protein language models (pLMs) attempt to learn these constraints directly from sequence~\citep{rives2021biological,elnaggar2021prottrans, nijkamp2023progen2, ferruz2022protgpt2}, typically using bidirectional Transformer~\citep{vaswani2017attention} encoders trained with masked language modeling (MLM)~\citep{devlin2019bert}. By predicting masked residues from their surrounding context, these models must capture essential interactions, including those that are far apart in sequence but close in $3$D space. The resulting conditional distributions enable zero-shot mutation scoring and contact recovery, while contextual representations enable downstream classification and regression tasks.

Recent progress in pLMs has largely been driven by scaling both model size and training data~\citep{esm2, nijkamp2023progen2,  esmc}. However, parameter count alone does not determine performance. AMPLIFY-$350$M~\citep{amplify}, despite having $43 \times$ fewer parameters than ESM-$2$~$15$B~\citep{esm2}, achieves comparable pseudo-perplexity on its curated validation set and better residue recovery on the human proteome. Several works further indicate that scaling can lead to undesirable memorization in larger pLMs~\citep{cheng2024training, amplify, hou2026understanding}. These observations motivate test-time scaling and adaptive computation as complementary axes to increase compute while keeping the number of parameters fixed.

\begin{figure}[t]
    \centering
    \includegraphics[width=0.75\textwidth]{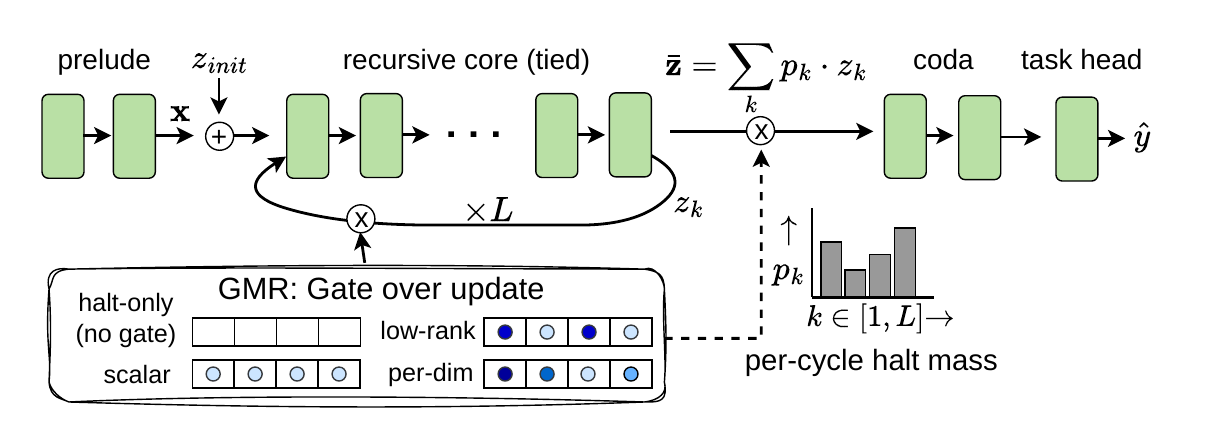}
   \caption{\textbf{recAMPLIFY architecture.} A non-recurrent prelude maps input embeddings to a residue-aligned latent state $z_k$, a weight-tied Transformer core iteratively refines this state, and a non-recurrent coda maps the refined state to the output space. Gated Mixture Recurrence (GMR) decodes from a probability-weighted mixture over the recurrence trajectory, with a per-sequence residual gate at scalar, low-rank, or per-dimension channel resolution controlling how much of each cycle's proposed update is accepted. Blue circles illustrate these channel-level parameterizations.}
    \label{fig:main_method_fig}
\end{figure}

Test-time compute scaling in autoregressive LLMs typically relies either on generating verbalized reasoning tokens~\citep{wei2022cot} or iteratively updating continuous latent states~\citep{ut, looped, hao2025training}. For masked pLMs with no autoregressive generation loop, the discrete-token approach does not transfer. Furthermore, amino-acid sequences do not admit verbalized rationales, and inserting thought tokens would break the residue-to-state alignment required for masked prediction. In contrast, iterative refinement of continuous latent representations, hereafter latent refinement, offers a natural way to increase pLM computation. Latent refinement already has a strong precedent in protein structure prediction, where AlphaFold~\citep{jumper2021alphafold} and ESMFold~\citep{esm2} recycle internal representations for a predefined number of cycles. However, these recycles operate on structural rather than sequence representations, leaving input-dependent test-time latent refinement unexplored in pLMs.

To approach test-time latent refinement, we hold a set of parameters as a recurrent unit~\citep{nakano1972associatron, amari1972learning} that iteratively refines its hidden states in a recursive fashion. Building on depth-recurrent language models~\citep{ut, looped, geiping2026scaling}, we adapt a recursive pretraining framework for test-time scaling in pLMs. Applied to AMPLIFY, this framework yields \emph{recursive AMPLIFY} (recAMPLIFY) that separates effective model depth from unique parameter count. In recAMPLIFY, a non-recurrent \emph{prelude} first maps token embeddings into a residue-aligned latent representation space where recursion can be unrolled. A weight-tied \emph{core} refines this state over a number of recurrent cycles, increasing compute without changing sequence length or residue alignment. A non-recurrent \emph{coda} then maps the refined state to the MLM output space for masked-residue prediction (Figure~\ref{fig:main_method_fig}). Because the core can be reapplied, the number of recurrent cycles, and therefore the effective depth, remains adjustable after pretraining.\footnote{We define \emph{depth recurrence} as the architectural mechanism of repeatedly applying a weight-tied block and \emph{recursive computation} as the resulting strategy of increasing effective depth through repeated latent refinement.}

Exposing recurrence depth as a test-time variable leaves open how to allocate the compute budget across inputs: fixed-depth recursion applies the same number of cycles to every input, yet we observe task- and assay-specific performance optima rather than monotonic gains from more cycles (Figures~\ref{fig:dms_bimodal_L} and~\ref{fig:vary_leval_sweep}). An effective recursion mechanism should thus decide both \emph{when} to stop the update and \emph{how} each cycle's update should shape the hidden state. Recent work in limited data regimes addresses the \emph{when} question by training halting heads from verifiable answers~\citep{trm}. However, no such verifier exists for MLM since determining which residues are biologically valid at a masked position is itself the problem the pLM is being trained to solve. Instead, we propose \emph{Gated Mixture Recurrence} (GMR), which decodes from a probability-weighted mixture of the hidden states visited during recursion and gates each cycle's residual update by a learned survival signal at a \emph{scalar}, \emph{low-rank}, or \emph{per-dimension} channel resolution. When the state update is ungated and the decode weights come from an independent halt head, we recover PonderNet-style halting~\citep{pondernet} as a special case of GMR. By unifying halt and gate into one framework, every learned GMR survival makes the recurrence trajectory itself input-adaptive, with the three gate resolutions progressing from coarser sequence-wide to finer feature-resolved control.

Our experiments on zero-shot and supervised probes show that recAMPLIFY with $155$M unique parameters, after being pretrained on the same $2$T tokens, remains competitive with the best performing AMPLIFY-$350$M  with $56$\% fewer unique parameters. Namely, recAMPLIFY improves fold accuracy by $22.5\%$ (from $56.8\%$ to $69.6\%$) and localization accuracy by $3.5$\% (from $71.6\%$ to $74.1\%$). It also outperforms the similar-sized AMPLIFY-$120$M across  supervised probes by up to $23$\%. We further observe that more recursion is not always better: recAMPLIFY trained with fixed recurrence depth can degrade beyond the recursion horizon seen during training. While variable-depth pretraining improves robustness to unseen depths, it does not provide a single optimal test-time depth. In contrast, GMR enables adaptive recursion while reducing the expected number of recurrent cycles by $32$--$40$\% relative to a recAMPLIFY trained with six cycles, while improving CASP contact recovery and most supervised probes at a small cost in zero-shot DMS and KL calibration. Our mechanistic analyses further show that residual gating, rather than halting alone, stabilizes recurrence beyond the trained horizon, while the expected halt depth reflects a sequence's composition entropy. Our main contributions are:

\begin{itemize}[leftmargin=*, itemsep=2pt, topsep=0pt, parsep=0pt, partopsep=0pt]
	\item We introduce a recursive pretraining framework for test-time scaling in masked pLMs by separating effective depth from unique parameter count.
    \item We propose recAMPLIFY, a recursive pLM with $56$\% fewer unique parameters than AMPLIFY-$350$M, and show its competitiveness on downstream tasks after training on the same $2$T tokens.
    \item To enable adaptive recursion in recAMPLIFY, we propose \emph{Gated Mixture Recurrence} (GMR), a unified framework that decodes from a mixture over recurrence-depth hidden states and gates their updates by a learned survival at scalar, low-rank, or per-dimension resolution.
	\item  We characterize the recurrent dynamics and the biological signal learned by recAMPLIFY, showing residual gating to stabilize recurrence and adaptive depth to track biological properties.
\end{itemize}

\section{Related Work}
\label{sec:related}

\paragraph{pLMs and iterative refinement.}

ESM-1b~\citep{rives2021biological}, ProtTrans~\citep{elnaggar2021prottrans}, and ESM-2~\citep{esm2} established that unsupervised protein representations discover aspects of structure, function, and variant-effect prediction. AMPLIFY~\citep{amplify} showed that data curation and architecture can matter more than parameter count. Iterative refinement is not new to protein modeling either. The folding trunks of AlphaFold~\citep{jumper2021alphafold} and ESMFold~\citep{esm2} repeatedly refine sequence and pair representations. However, these recycles belong to a supervised structure-prediction trunk and use a globally chosen count. Instead, we pretrain with recurrence inside the sequence encoder, and study both fixed and input-dependent test-time depth.

\paragraph{Scaling beyond parameters.}
\label{sec:scaling_beyond}

Popular methods for scaling language models with fixed parameter count include repeated sampling~\citep{cobbe2021gsm8k, baek2026generative}, chain-of-thought (CoT) reasoning~\citep{wei2022cot}, or both~\citep{wang2023selfconsistency, yoran2023answering}. Nevertheless, these come with their own trade-offs. Repeated sampling needs additional verifiers  for  assessing candidates, which is not always possible for scientific discovery tasks~\citep{cornelio2025need}. CoT reasoning calls for committing to discrete verbalized tokens. Addressing these pitfalls, a third scaling axis involves thinking recursively in continuous latent spaces~\citep{amari1972learning,  hao2025training}. Universal~\citep{ut} and Looped Transformers~\citep{looped} approach this through a weight-tied, iterative refinement lineage ~\citep{liao2016bridging}. Given the effectiveness of recursive thinking for limited data regimes~\citep{hrm, trm, srm} and how it helps compact models with better learning priors~\citep{NEURIPS2021_3501672e, bansal2022end, kohli2026loop}, we hypothesize that weight-tied recurrence is crucial for scaling compact pLMs, where larger models are known to memorize training data~\citep{amplify}. A desirable property for such recurrence  is adaptive computation~\citep{act} that  refines the latent state in an input-dependent manner~\citep{blayney2026mechanistic, fein2026solve}.  Here, a halting head can be trained either in a supervised~\citep{trm} or self-supervised manner~\citep{pondernet} to recognize convergence. For a pLM encoder, supervised halting requires a verifier to recognize which residues are valid at each masked position, which is exactly the biology the model is being trained to learn. We thus adopt the latter self-supervised objective.

\paragraph{Mechanistic analyses of recursion.}

Depth recurrence defines a learned dynamical system whose hidden states can converge or oscillate, and where additional iterations can degrade a previously useful representation. Prior work links successful depth extrapolation to input recall~\citep{bansal2022end}, path-independent attractors~\citep{anil2022path, geiping2026scaling}, and stable cyclic fixed points~\citep{prairie2026parcae}, and identifies ``overthinking'' beyond the trained depth as a recurring failure mode~\citep{kohli2026loop,blayney2026mechanistic}. These studies treat recursion dynamics, such as depth sweeps, state changes, output convergence, and evolutionary properties, as operational rather than interpretive. We build on these works in our analysis of gated recurrence and protein properties.

\section{Problem Setup}
\label{sec:problem}

\paragraph{Test-time scaling through recursive pretraining.}

We formulate test-time scaling for masked pLMs using AMPLIFY~\citep{amplify}, whose released pretraining data, code and architecture recipe enable controlled comparisons under a fixed data budget. Let $\seq=(\resi[1],\ldots,\resi[\seqlen])$ denote a protein sequence of length $\seqlen$, where $\resi$ denotes the residue at position $i$, drawn from the alphabet of 20 standard amino acids. AMPLIFY pretrains a pre-norm Transformer encoder $f$ by minimizing the masked-token cross-entropy $\Lmlm$ to recover masked residues from a bidirectional context. Its hidden states remain in one-to-one alignment with the input residues, supporting masked-marginal mutation scoring and frozen-embedding probes. We seek to build a recursive AMPLIFY encoder whose inference-time computation can be increased per protein without changing its unique parameter count or disrupting residue alignment. To this end, we expose the number $\ncyc$ of applications of a weight-tied Transformer core as a test-time variable. This allows the model to iteratively refine its residue-aligned latent state under varying compute budgets.

\begin{wrapfigure}{r}{0.35\textwidth}
    \vspace{-0.2in}
    \centering
    \includegraphics[width=0.9\linewidth]{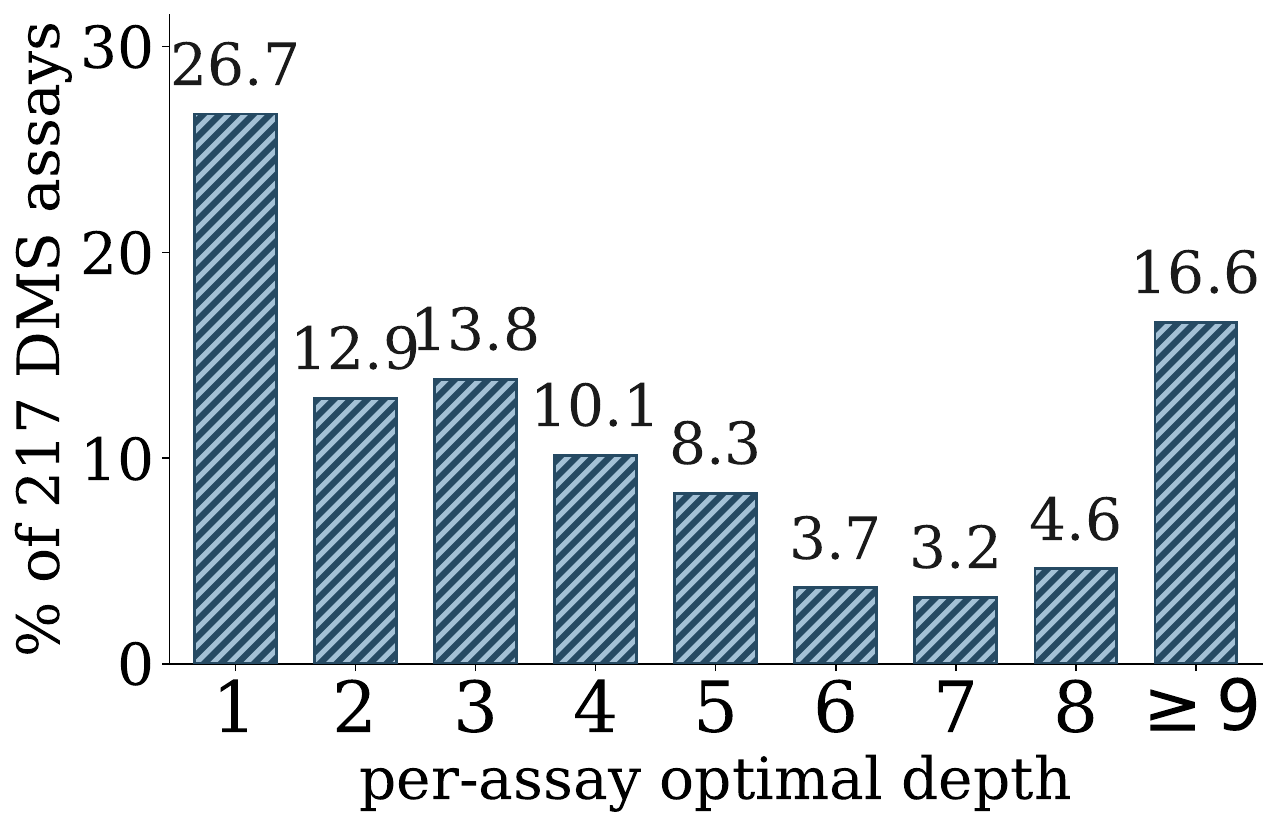}
    \vspace{-0.1in}
    \caption{recAMPLIFY's per-assay optimal recurrence depth on ProteinGym DMS.}
    \label{fig:dms_bimodal_L}
\end{wrapfigure}

\paragraph{Desired properties of a recursive pLM.}
\label{para:whatweask}

Proteins vary widely in biological complexity, from short signal peptides to multi-domain enzymes and from structured domains to intrinsically disordered chains. pLMs must accommodate diverse proteins, and the per-assay optima suggest that no single recurrence depth is optimal across all proteins (Figure~\ref{fig:dms_bimodal_L}). These observations motivate learning a recurrence depth budget that adapts to each sequence. We thus seek a recursive pLM that can allocate compute adaptively while preserving stable latent dynamics. We then study whether its expected halt depth carries interpretable information about proteins.

\section{Method}
\label{sec:method}

We start by adapting the minimal recurrent design of \citet{geiping2026scaling} from a causal decoder to the bidirectional AMPLIFY encoder $f$, yielding recAMPLIFY. Here, the early $p$ layers of the Transformer encoder form the \emph{prelude} that runs once to produce the input representation $\emb \in\mathbb{R}^{\seqlen\times\dmodel}$ from token embeddings. This is followed by an $r$-layer weight-tied \emph{recurrent core} that repeatedly refines a single state $\zlat\in\mathbb{R}^{\seqlen\times\dmodel}$ over $\ncyc$ cycles before passing it to a $c$-layer \emph{coda} that decodes the final state $\widetilde{\zlat}$ from the latest cycle: $\widetilde{\zlat} = \zstep[\ncyc]$. For the core to act as a stable data-dependent iterative operator, we re-inject the input $\emb$ at every cycle~\citep{bansal2022end}. To encourage a more general solution, we initialize the latent state with a fixed vector shared by all residues, $\zinit = \mathbf{1}_{\seqlen},\mathbf{u}^{\top}$, where $\mathbf{u}\in\mathbb{R}^{\dmodel}$ is drawn once from $\mathcal{N}(0,\sigma^2 I)$, with $\sigma = 0.02$ at model initialization and kept frozen, so every forward pass starts from the same state \citep{anil2022path}:
\begin{equation}
    \emb=\preludefn(\Embed(\seq)),\qquad
    \zstep[\kcyc]=\fcore \left(\zstep[\kcyc-1],\emb\right),\qquad
    \pred=\text{LMHead}(\codafn \left(\widetilde{\zlat}\right)),
    \label{eq:minimal_recurrence}
\end{equation}
where $\kcyc\in\{1,\ldots,\ncyc\}$ is the recurrent cycle. The coda output is passed to the LM head for masked-token prediction and zero-shot probing. For supervised evaluation, we train a two-layer MLP on frozen representations: the recurrent-core state for fixed- and variable-depth recAMPLIFY, and the coda output for GMR variants (the two choices are compared in Figure~\ref{fig:vary_embed_mode}). We use the triplet $(p,r,c)$ for the unique layer counts and $\ncyc$ for the number of cycles. This gives an effective depth of $p+(r\cdot\ncyc)+c$ while exposing $\ncyc$ as a globally specified inference-time compute value. Motivated by the heterogeneous depth need in \S\ref{para:whatweask}, we next make this budget input-dependent.

\subsection{Gated Mixture Recurrence}
\label{sec:gmr}

Adaptive recursion requires allocating compute along two axes: \emph{when} recurrent refinement should stop, and \emph{how} each cycle's update should shape the hidden state along the way. We propose \emph{Gated Mixture Recurrence} (GMR), a framework that treats both these axes under a single learned survival signal aimed at unifying halting and gating. The recurrent core produces a trajectory of intermediate hidden states $\{\zstep[k]\}_{k=1}^{\ncyc}$. GMR decodes each input from a probability-weighted mixture $\bar{\zlat}$ while gating each state update $\zstep[k]=\zstep[k-1]+S_k\odot(\widetilde{\zlat}_k-\zstep[k-1])$ by a channel-resolved survival $S_k$.

\paragraph{When to stop updating?}

As a supervised adaptive computation head~\citep{trm} would require a per-position verifier (\S\ref{sec:scaling_beyond}), we instead adopt a self-supervised approach~\citep{pondernet} within GMR. Here, a halt head reads the state and emits a conditional halt probability $\lam[k]=\texttt{sigmoid}(\halthead(\stopgrad\,\widetilde{\zlat}_{k}))$, where the stopgrad operator $\stopgrad$ prevents the head's gradients from directly steering the LM features. This yields the unconditional halt mass $\phalt[k]$ that acts as mixture weights, with the training loss $\mathcal{L}$ additionally maximizing the entropy $H$ to encourage more cycles:
\begin{equation}
    \phalt[k] = \lam[k]\textstyle\prod_{j<k}(1-\lam[j]),
    \qquad
    \bar{\zlat}=\sum_k\phalt[k]\,\zstep[k],
    \qquad
    \Loss = \Lmlm(\text{LMHead}(\codafn(\bar{\zlat}))) - H(\{\phalt\}),
    \label{eq:ponder}
\end{equation}
where $\zstep[k]$ is the $k$-th cycle output. We decode predictions from the mass-weighted state $\bar{\zlat}$ and stop once the continuation mass $\prod_{j<k}(1-\lam[j])$ falls below a threshold $\varepsilon$ and discard this residual mass. Since the continuation mass is non-increasing, probability committed to halting at cycle $j$ cannot be redistributed to later cycles.

\paragraph{How to update?}
\label{sec:gates}

In the halt-only case described above, a single halt probability weights the decode mixture but leaves state updates ungated, treating all hidden-state features as converging together. We relax this by allowing survival to vary across $c$ channels of the feature space, so different features can stabilize at different cycles. To this end, a residual gate reads the proposal $\widetilde{\zlat}_{\kcyc}$ and emits channel-wise values $g_{\kcyc}^{\text{ch}}=\texttt{sigmoid}(\halthead\,(\stopgrad\,\widetilde{\zlat}_{\kcyc}))$, where $g^{\text{ch}} \approx 1$ preserves the channel's future update potential and $g^{\text{ch}} \approx 0$ suppresses subsequent updates. Gate values accumulate across cycles into the survival $S_{\kcyc}^{\text{ch}}$, which is broadcast or projected to the full hidden dimension $D$ to yield the per-feature update weights $S_{\kcyc}^{D}$. Put together, the gated state update is:
\begin{equation}
    \zstep[\kcyc] = \zstep[\kcyc-1] + S_{\kcyc}^{D}\odot\big(\widetilde{\zlat}_{\kcyc}-\zstep[\kcyc-1]\big),
    \qquad S_{\kcyc}^{D} = \texttt{project}(S_{\kcyc}^{\text{ch}}), \qquad
    S_{\kcyc}^{\text{ch}}=\prod_{j<\kcyc}g_j^{\text{ch}},
    \label{eq:gate}
\end{equation}
where $S_{\kcyc}^{\text{ch}}$ is the update magnitude at cycle $\kcyc$ set by the accumulated survival of previous gates, while the newly predicted gate $g_{\kcyc}^{\text{ch}}$ contributes only to the current cycle's halt mass. Since the survival $S_{\kcyc}^{\text{ch}}$ is monotonically non-increasing, this inherits the halt semantics of equation~\ref{eq:ponder}, where once a channel has committed to halting, its future updates are downweighted.

We instantiate the gate at three channel-level granularities: (a) the \emph{scalar} gate broadcasts a single survival value (${\text{ch}}=1$) across all feature channels. This is similar to the halt scalar $\lam$ from equation~\ref{eq:ponder}, but acts on state-update magnitude rather than on decode; (b) the \emph{low-rank} gate maintains ${\text{ch}} \ll \dmodel$ shared survival trajectories (e.g., ${\text{ch}} = 16$, $\dmodel = 960$) and projects them onto every hidden feature through a learned row-stochastic assignment. This ensures that all $\dmodel$ features are steered by the same shared survivals; (c) the \emph{per-dimension} gate is our most granular variant, using one independently predicted survival trajectory per channel (${\text{ch}} = \dmodel$). Halting occurs either when we reach the predefined limit $\ncyc$ or when every channel has a survival below the threshold $\varepsilon$ (see Listing~\ref{alg:ar} for the pseudocode).

\section{Experiments}
\label{sec:experiments}
\paragraph{Training and evaluation.}

Following AMPLIFY’s pretraining recipe and using its publicly released code and data~\citep{amplify}, we pretrain recAMPLIFY on the same mixture of the UniRef100~\citep{10.1093/bioinformatics/btm098}, SCOP~\citep{murzin1995scop}, and OAS~\citep{olsen2022observed} databases, making the original AMPLIFY our direct baseline. Further details are provided in \S\ref{app:pretraining_setup}.

Following \citet{esm2}, we evaluate using two types of probes. \emph{Zero-shot probes:} KL calibration~\citep{LeBreton2026}, ProteinGym DMS~\citep{proteingym}, and CASP16 contact recovery~\citep{yuan2026casp16}. \emph{Supervised probes:} we use the $18$-task benchmark of \citet{chen2024xtrimopglm} and study four representative tasks reported to show clear benefits from pretraining~\citep{LeBreton2026}: fold accuracy~\citep{rao2019evaluating}, localization accuracy~\citep{dallago2021flip}, stability prediction~\citep{rao2019evaluating, dallago2021flip}, and secondary-structure prediction (SSP)~\citep{rao2019evaluating}. For each supervised task, we report the mean test performance over $5$ seeds after re-training the best configuration from a $50$-trial Bayesian TPE hyperparameter search. Further evaluation and hyperparameter details are provided in \S\ref{app:eval_details}.

\begin{table}[t]
    \centering
    \caption{\textbf{Zero-shot and supervised performance.} The top block reports ESM-2 for context, as its pretraining data favor structural modeling while AMPLIFY’s data emphasize evolutionary calibration on consensus positions (KL). The middle block reports full-scale comparisons, where recurrence leads on CASP, stays close on KL (NT) and DMS, and improves three out of four supervised probes. The bottom block reports ablations with variable-depth training and four GMR variants. Effective depth $\effdepth=p + \ncyc r + c$ counts layer applications, using $\ncyc_{\mathrm{eval}}$ at inference and expected cycles for GMR. Best scores within each block are \textbf{bold}. See Table~\ref{tab:srm_vs_baseline_1m_std} for standard deviations.}
    \label{tab:srm_vs_baseline_1m}

    \begin{adjustbox}{max width=\linewidth}
        \begin{tabular}{lccccccccccc}
            \toprule
            & \multicolumn{4}{c}{\textbf{Zero-shot probes}}
            & \multicolumn{4}{c}{\textbf{Supervised probes}}
            & \multicolumn{3}{c}{\textbf{Param / compute}} \\
            \cmidrule(lr){2-5} \cmidrule(lr){6-9} \cmidrule(lr){10-12}
            Model
            & KL (T)\,$\uparrow$
            & KL (NT)\,$\uparrow$
            & DMS $\rho$\,$\uparrow$
            & CASP P@$\seqlen$\,$\uparrow$
            & Fold acc.\,$\uparrow$
            & Loc.\ acc.\,$\uparrow$
            & Stability $\rho$\,$\uparrow$
            & SSP (Q$3$+Q$8$)\,$\uparrow$
            & $N$
            & $\effdepth^{\mathrm{train}}$
            & $\effdepth^{\mathrm{eval}}$ \\
            \midrule
            \multicolumn{12}{c}{\textbf{ESM-2 baselines for context (}$\mathbf{500}$\textbf{K pretraining steps on UniRef50/90})} \\
            \midrule
            ESM-2-$150$M & $37.79$ & $31.54$ & $0.388$ & $0.239$ & $0.645$ & $0.734$ & $\best{0.764}$ & $0.777$ & $149$M & $30$ & $30$ \\
            ESM-2-$650$M & $55.07$ & $41.51$ & $\best{0.415}$ & $0.268$ & $0.675$ & $0.759$ & $0.659$ & $0.805$ & $652$M & $33$ & $33$ \\
            ESM-2-$3$B & $\best{63.11}$ & $\best{44.22}$ & $0.407$ & $\best{0.281}$ & $\best{0.723}$ & $\best{0.782}$ & $0.694$ & $\best{0.808}$ & $2.8$B & $36$ & $36$ \\
            \midrule
            \multicolumn{12}{c}{$\mathbf{1}$\textbf{M pretraining steps on UniRef100}} \\
            \midrule
            AMPLIFY-$120$M & $57.33$ & $37.87$ & $0.345$ & $0.126$ & $0.566$ & $0.651$ & $0.591$ & $0.724$ & $118$M & $24$ & $24$ \\
            AMPLIFY-$350$M & $\best{69.31}$ & $\best{38.79}$ & $\best{0.360}$ & $0.146$ & $0.568$ & $0.716$ & $0.585$ & $\best{0.742}$ & $354$M & $32$ & $32$ \\
            fixed-$L$ recAMPLIFY & $59.32$ & $37.70$ & $0.348$ & $\best{0.179}$ & $\best{0.696}$ & $\best{0.741}$ & $\best{0.703}$ & $\best{0.742}$ & $155$M & $44$ & $44$ \\
            \midrule
            \multicolumn{12}{c}{$\mathbf{100}$\textbf{K pretraining steps on UniRef100}} \\
            \midrule
            AMPLIFY-$120$M & $49.84$ & $36.07$ & $0.345$ & $0.120$ & $0.602$ & $0.691$ & $0.611$ & $0.720$ & $118$M & $24$ & $24$ \\
            AMPLIFY-$350$M & $\best{59.79}$ & $\best{38.79}$ & $\best{0.359}$ & $0.104$ & $0.570$ & $0.672$ & $0.479$ & $\best{0.729}$ & $354$M & $32$ & $32$ \\
            vary-$\ncyc$ recAMPLIFY & $48.88$ & $36.22$ & $0.352$ & $0.128$ & $\best{0.653}$ & $0.711$ & $0.688$ & $0.726$ & $133$M & $52$ & $52$ \\
            fixed-$\ncyc$ recAMPLIFY & $49.92$ & $36.59$ & $\best{0.359}$ & $0.117$ & $0.601$ & $0.700$ & $0.644$ & $0.720$ & $133$M & $52$ & $52$ \\
            $+$ halt-only GMR & $49.71$ & $36.27$ & $0.347$ & $0.135$ & $0.637$ & $0.714$ & $0.650$ & $\best{0.729}$ & $133$M & $52$ & $33$ \\
            $+$ scalar gated GMR & $49.82$ & $36.45$ & $0.346$ & $\best{0.161}$ & $0.628$ & $0.708$ & $\best{0.737}$ & $0.722$ & $133$M & $52$ & $37$ \\
            $+$ rank-$16$ gated GMR & $50.39$ & $36.51$ & $0.346$ & $0.150$ & $0.621$ & $0.708$ & $0.703$ & $0.721$ & $133$M & $52$ & $36$ \\
            $+$ per-dim gated GMR & $50.36$ & $36.45$ & $0.347$ & $0.143$ & $0.621$ & $\best{0.718}$ & $0.627$ & $0.722$ & $133$M & $52$ & $37$ \\
            \bottomrule
        \end{tabular}
    \end{adjustbox}
\end{table}

\begin{figure}[t]
    \centering
    \begin{subfigure}[t]{0.38\textwidth}
        \centering
        \includegraphics[width=0.95\linewidth]{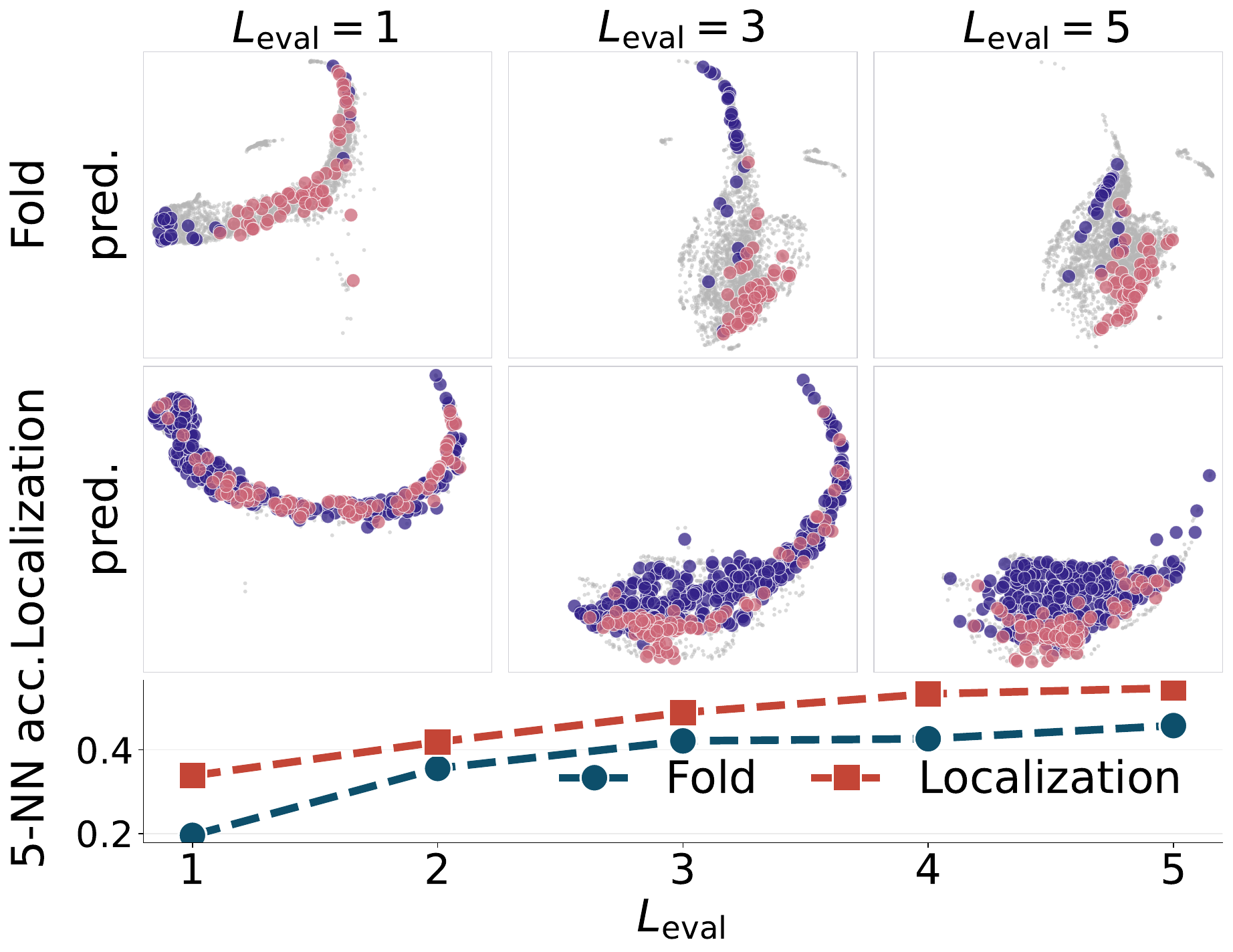}
        \caption{UMAP of sequence embeddings.}
        \label{fig:vary_leval_umap}
    \end{subfigure}\hfill
    \begin{subfigure}[t]{0.62\textwidth}
        \centering
        \includegraphics[width=0.99\linewidth]{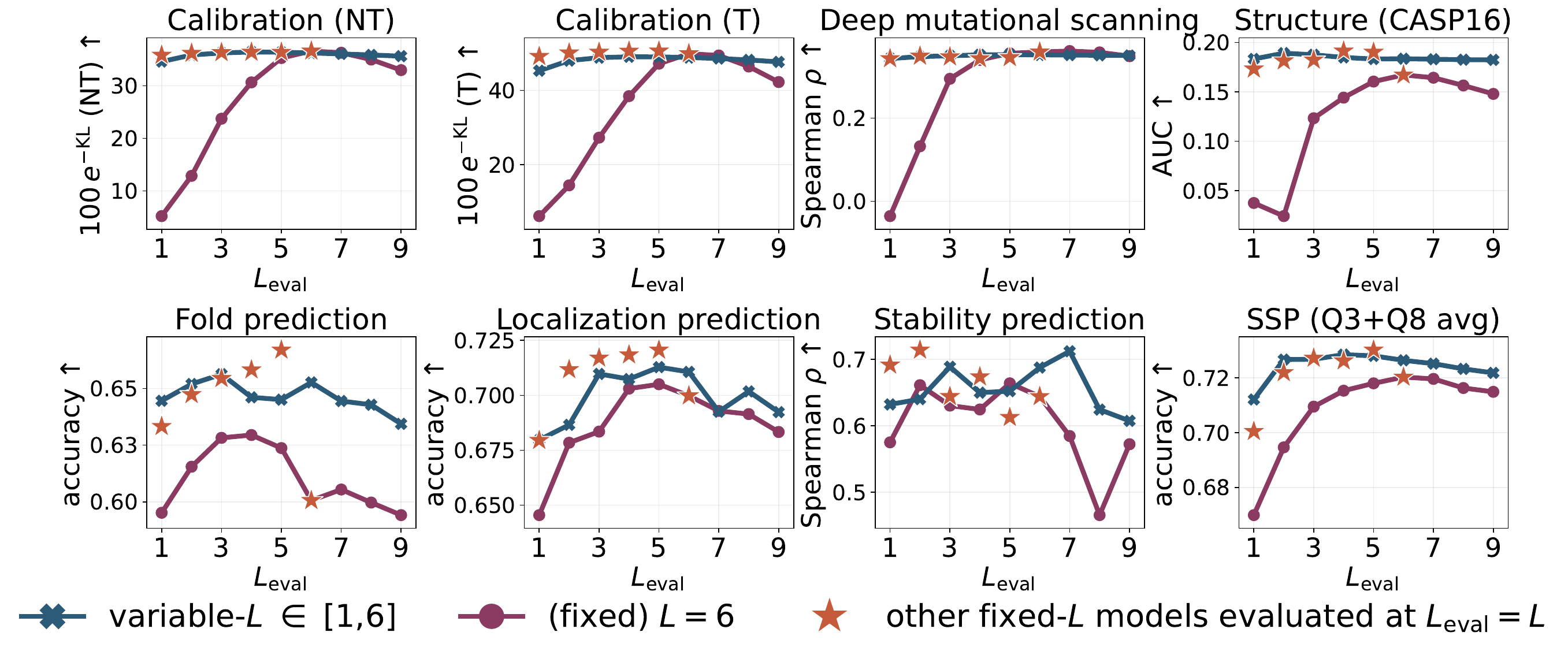}
        \caption{Test-time $\ncyc_{\mathrm{eval}}$ sweep across $100$K pretraining setups.}
        \label{fig:vary_leval_sweep}
    \end{subfigure}
    \caption{\textbf{recAMPLIFY at varying recurrence depths.} (a) For $1$M-step recAMPLIFY, the coda embeddings become progressively more class-separable ($5$-NN accuracy) with larger test-time depth $\ncyc_{\mathrm{eval}}$. (b) For $100$K-step recAMPLIFY, as $\ncyc_{\mathrm{eval}}$ increases past the trained depth ($\ncyc = 6$), the fixed-$\ncyc$ model (red) degrades while the variable-$\ncyc$ variant (blue) remains largely stable.}
    \label{fig:vary_leval}
\end{figure}

\subsection{Scaling and Depth Behavior}

\begin{figure}[t]
    \centering
    \begin{subfigure}[t]{0.42\textwidth}
        \centering
        \includegraphics[width=0.9\linewidth]{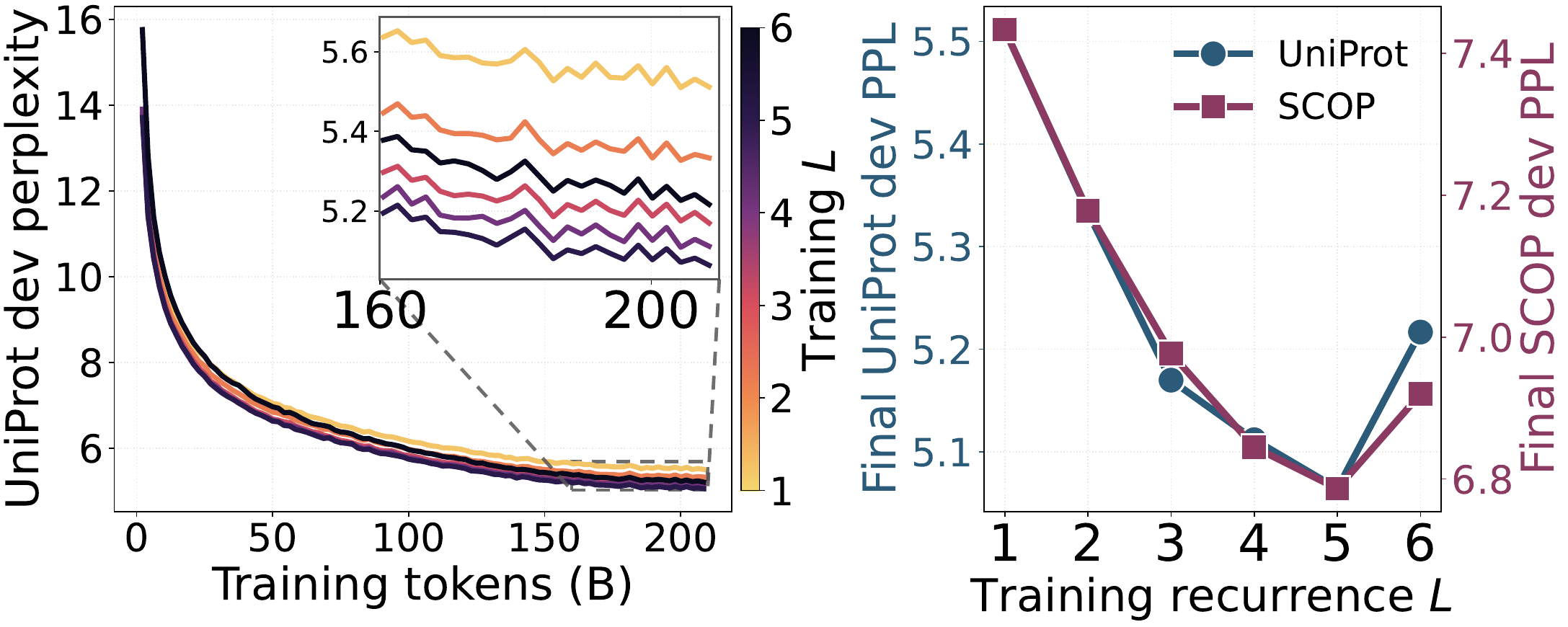}
        \caption{Held-out set perplexity.}
        \label{app:ppl_devset}
    \end{subfigure}\hfill
    \begin{subfigure}[t]{0.27\textwidth}
        \centering
        \includegraphics[width=0.9\linewidth]{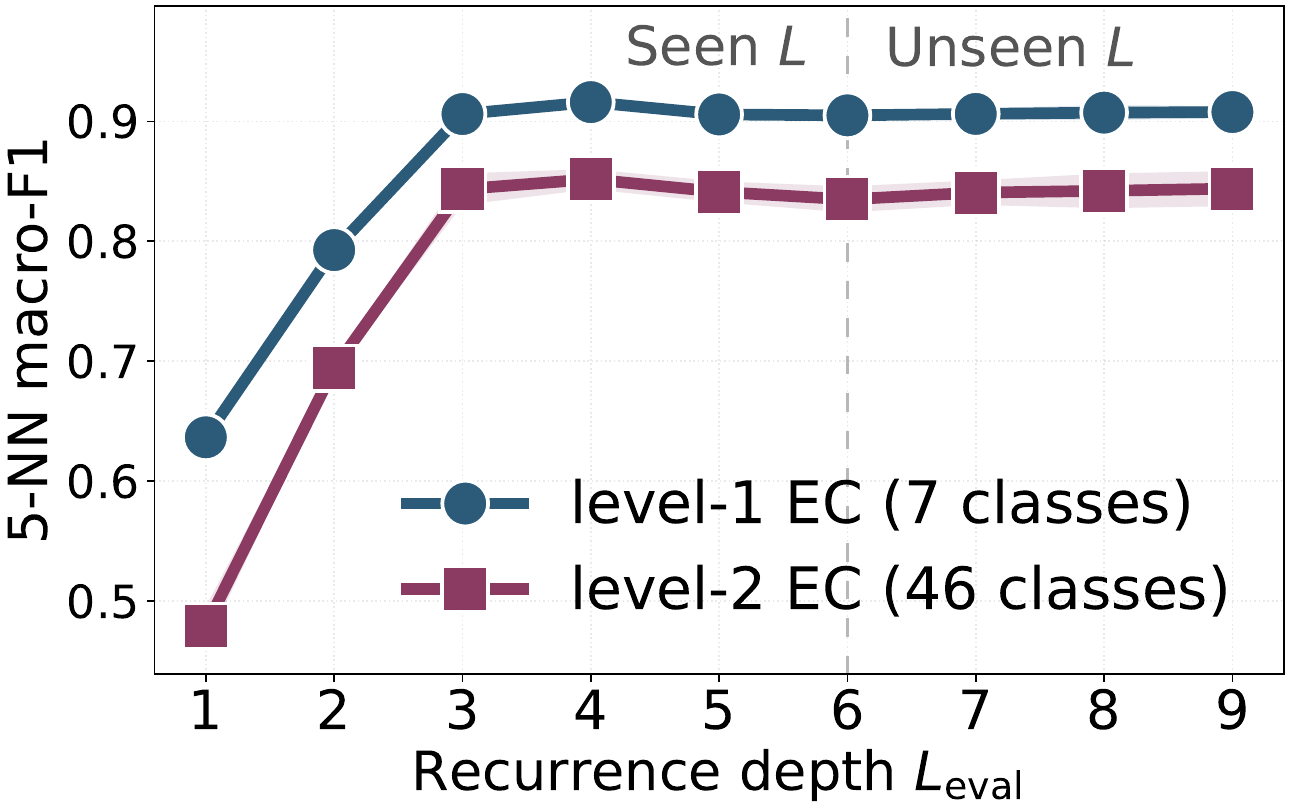}
        \caption{Enzyme classification.}
        \label{app:ec_knn_leval}
    \end{subfigure}\hfill
    \begin{subfigure}[t]{0.29\textwidth}
        \centering
        \includegraphics[width=0.9\linewidth]{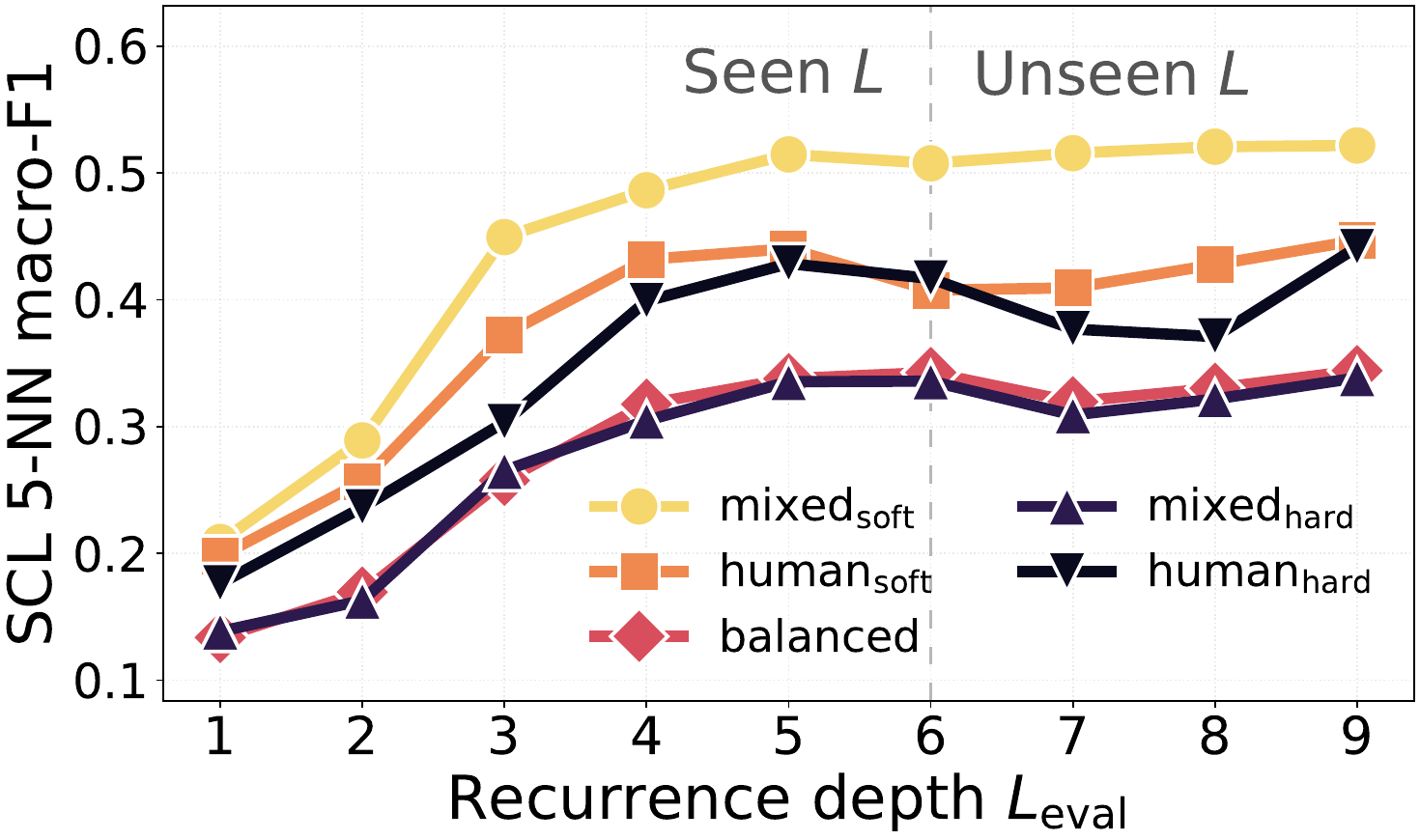}
        \caption{Subcellular localization.}
        \label{app:subcellular_loc}
    \end{subfigure}
    \caption{\textbf{Effect of recurrence depth on sequence likelihood and downstream separability at $\mathbf{100}$K steps.} (a) UniProt perplexity for fixed-$L$ pretraining, evaluated at $\ncyc_{\mathrm{eval}} = L$. (b) EC level-1 and level-2 $5$-NN macro-F1 across $\ncyc_{\mathrm{eval}}$ for recAMPLIFY trained with $\ncyc = 6$, using $14$K Swiss-Prot enzymes balanced across the $7$ top-level classes relabeled into $46$ level-2 subclasses ($ \geq 30$ samples each). (c) SCL $10$-way $5$-NN macro-F1 across five FLIP DeepLoc-1.0 splits~\citep{didi2026flip2}, fitted on $11{,}181$ shared training sequences and evaluated on each held-out test set.}
    \label{fig:depth_vs_perf}
\end{figure}

\paragraph{Does recursive pretraining scale?}

We first examine whether the minimal recurrent design of recAMPLIFY remains viable at the full pLM pretraining scale of $1$M steps. For a fair comparison, we match AMPLIFY-$350$M's $15$ attention heads and a hidden dimension of $960$, and set the number of layers to $(p,r,c) = (2,10,2)$, i.e., roughly the budget of the compact AMPLIFY-$120$M baseline. Table~\ref{tab:srm_vs_baseline_1m} shows that recAMPLIFY, with $155$M unique parameters, remains competitive with AMPLIFY-$350$M across downstream probes when both are trained on the same $2$T tokens. The clearest gains appear on the four supervised probe tasks that most reliably measure information acquired during pretraining. Visualizing the coda embeddings across recurrent cycles (Figure~\ref{fig:vary_leval_umap}) shows that each additional cycle brings the embedding manifold closer to the downstream label geometry, improving $5$-NN accuracy. Finally, we verify that the cost of our recursive pretraining recipe scales with depth alone. Table~\ref{tab:cost} shows that our $133$M recAMPLIFY at $L=3$ achieves measured effective throughput comparable to AMPLIFY-$350$M, which has approximately $2.7\times$ as many unique parameters.

\paragraph{Comparison with ESM-2.}

A direct comparison with ESM-2 is confounded by differences in model size and pretraining data: its larger checkpoints have $4$--$18\times$ as many parameters as recAMPLIFY and use an earlier UniRef release with a different clustering scheme. ESM-2 samples uniformly over UniRef50 clusters, whereas AMPLIFY’s UniRef100 recipe retains more of the observed sequence redundancy, resulting in different training distributions (\S\ref{app:pretraining_data_choice}). ESM-2’s unreleased pretraining code and exact data splits also make reproducing its training setup difficult. With these caveats, larger ESM-2 checkpoints outperform recAMPLIFY on three out of four supervised probes, while recAMPLIFY exceeds ESM-2-$650$M and $3$B on stability prediction and achieves better target (T) KL calibration than ESM-2-$150$M and $650$M.

\paragraph{A $\mathbf{100}$K-step regime for controlled ablations.}

Training recAMPLIFY for $1$M steps makes extensive ablations computationally impractical. Following \citet{LeBreton2026}, we next ask whether recursive pretraining at a smaller scale remains representative. Table~\ref{tab:srm_vs_baseline_1m} shows that a similar recAMPLIFY configuration, $(p,r,c)=(2,8,2)$, achieves a representative task-performance profile at $100$K steps. We therefore use $100$K steps for subsequent experiments.

\paragraph{How does recursion behave with depth?}

Figure~\ref{fig:depth_vs_perf} shows that greater depth improves both validation perplexity and downstream separability for enzyme classes (EC) and subcellular localization (SCL). For SCL, deeper recursion yields macro-F1 gains of $20$--$31$~pp across FLIP's train/test cluster splits. Given the diverse criteria used to construct these splits, this suggests benefits from recursion under distribution shift. Figure~\ref{fig:vary_leval_sweep} shows downstream probe performance as recAMPLIFY is trained with larger cycles $\ncyc\in\{1,\ldots,6\}$ (marked by stars). For the model trained with $\ncyc=6$, further varying $\ncyc_{\mathrm{eval}}$ shows that $\ncyc_{\mathrm{eval}}=\ncyc$ performs best on KL calibration, CASP and SSP, while Fold, Localization and Stability peak slightly earlier, $\ncyc_{\mathrm{eval}}\in\{4,5\}$. Finally, at a fixed depth, we find the physical layer counts $r=10$ (Figure~\ref{fig:vary_r}) and $(p,c)=(2,2)$ (Figure~\ref{fig:vary_prelude_coda}) to perform best overall.

\paragraph{Going beyond fixed-depth recursion.}

The fixed-depth recAMPLIFY baseline uses $\ncyc$ cycles during training and exposes recurrence depth as a test-time variable $\ncyc_{\mathrm{eval}}$. Figure~\ref{fig:vary_leval_sweep} shows that this model (red lines) degrades at unseen depths ($\ncyc_{\mathrm{eval}}>6$), most on stability ($0.644 \to 0.465$) and on zero-shot KL and CASP, with milder losses on the other supervised probes. In contrast, pretraining with a variable number of cycles sampled uniformly from $\{1,\ldots,6\}$ per minibatch generally improves robustness to unseen depths (blue lines). Several probes peak near the variable-depth model’s mean training depth, suggesting a simple heuristic: vary the cycle count during training, then fix $\ncyc_{\mathrm{eval}}$ near that mean. However, the per-assay optima in Figure~\ref{fig:dms_bimodal_L} suggest that no single depth is optimal across all assays. Put together, these observations support our motivation for an adaptive recursion mechanism that predicts when further refinement is no longer useful.

\paragraph{Gated Mixture Recurrence.}

Table~\ref{tab:srm_vs_baseline_1m} shows that every GMR variant reduces the \emph{expected} test-time depth from $52$ to $33$--$37$ layer applications while improving over fixed-$\ncyc$ recAMPLIFY on CASP contact recovery ($0.135$--$0.161$ vs.\ $0.117$), Fold, Localization and SSP, and on Stability for three of the four variants. The cost is a small drop in zero-shot DMS ($0.346$--$0.347$ vs.\ $0.359$) and KL calibration (at most $0.32$ NT points). We quantify depth as the mean of the model's predicted halting distribution: $\En=\sum_k k\,p_k$. This characterizes the learned allocation policy rather than measured wall-clock savings, since our evaluation decodes the full mixture $\sum_k p_k\mathbf{z}_k$ over the cycle budget. Gate resolution has no consistent effect on zero-shot calibration (KL (T) $49.7$--$50.4$ across variants), but gated variants lead the $100$K block on CASP and Stability (scalar) and on Localization (per-dim). Halt-only GMR reaches the shallowest expected depth ($\En=3.57$, i.e., $2+3.57{\times}8+2\approx33$ layers) while matching AMPLIFY-$350$M on SSP.  To understand where gating helps, we next examine whether the gates produce an empirical fixed point of the recurrent state (\S\ref{sec:gate_convergence}); whether recursion preserves the depth-heterogeneous layerwise mixing profile of a feedforward Transformer (\S\ref{sec:stages_of_inference}); and what the GMR halt signal encodes about proteins (\S\ref{sec:halt_biology}).

\subsubsection{Hidden State Evolution}
\label{sec:gate_convergence}

Let $\mathbf{z}^{\ell}_{\kcyc}$ denote the hidden state at physical layer $\ell \in \{1,\ldots,r\}$ of the recurrent block during cycle $\kcyc \in \{1,\ldots,\ncyc_{\mathrm{eval}}\}$. The block-level state $\zstep[\kcyc]$ used in \S\ref{sec:gmr} is its top-layer output, i.e., $\zstep[\kcyc] \equiv \mathbf{z}^{r}_{\kcyc}$. We analyze how recursion affects hidden-state dynamics within and beyond the depth seen during training. Following \citet{blayney2026mechanistic}, we unroll each recAMPLIFY variant to $\ncyc_{\mathrm{eval}}=50$ and examine state changes across recurrent layers along three complementary axes: (a) the size of successive updates $\|\Delta\zlat\|_2$, (b) the normalized distance of each cycle’s state to its deepest recorded state $\mathbf{z}^{\ell}_{\ncyc_{\mathrm{eval}}}$, taken as an approximate fixed point, and (c) the induced attention geometry, measured by pairwise Frobenius distances between attention maps across cycles.

\begin{figure}[t]
    \centering
    \begin{subfigure}[t]{0.49\textwidth}
        \centering
        \includegraphics[width=0.9\linewidth]{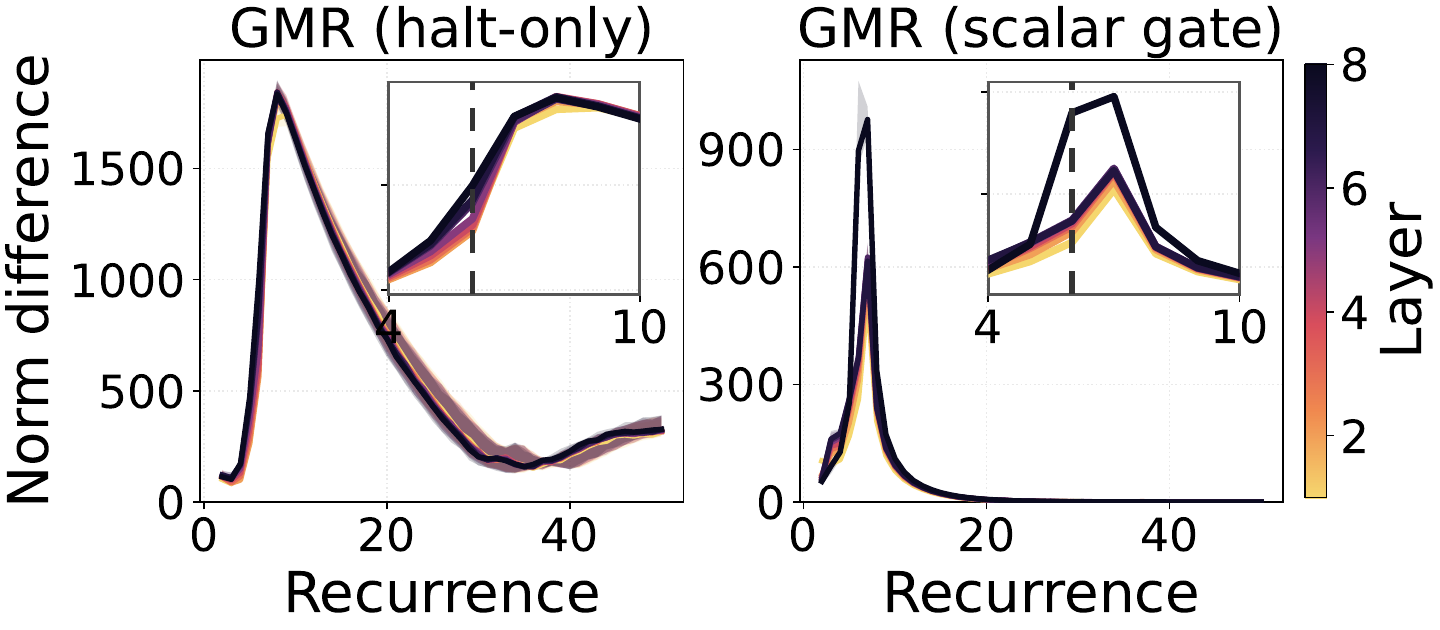}
        \caption{Difference across successive iterations.}
    \end{subfigure}
    \hfill
    \begin{subfigure}[t]{0.49\textwidth}
        \centering
        \includegraphics[width=0.9\linewidth]{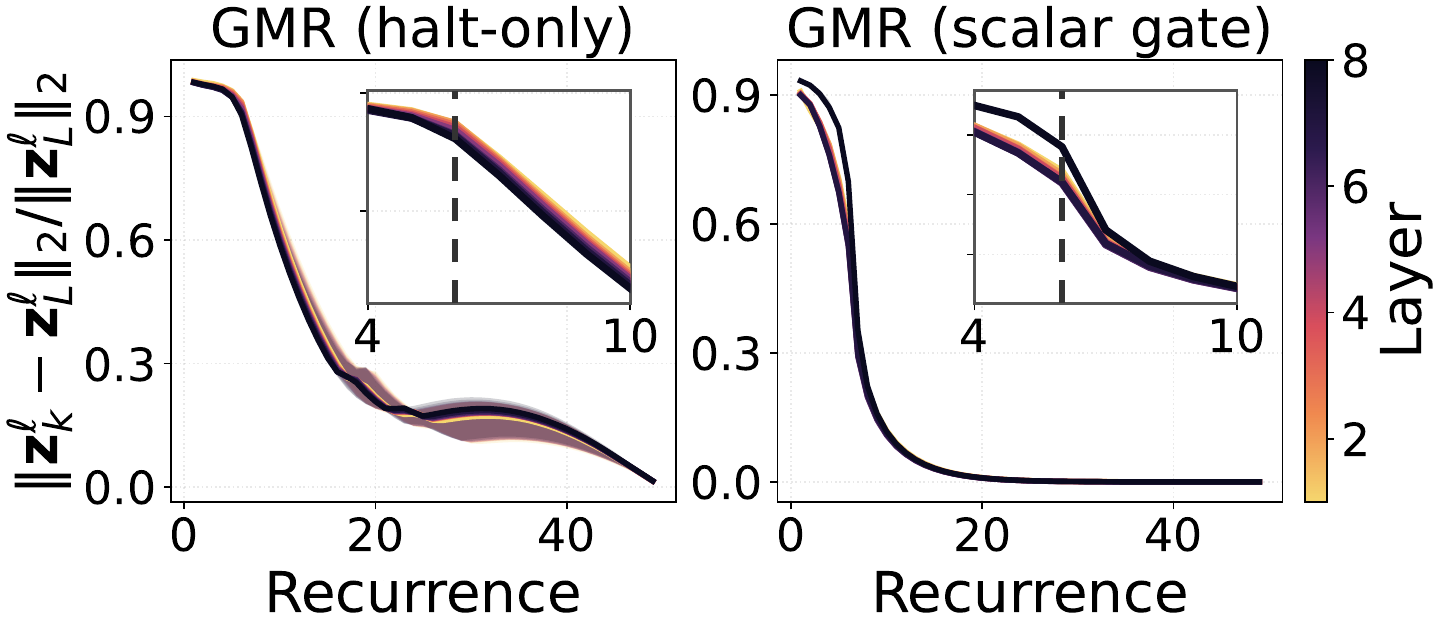}
        \caption{Normalized distance to the deepest state.}
    \end{subfigure}
   \caption{\textbf{Dynamics of recurrent block layer} $\ell$ on stability prediction upon training with $\ncyc=6$ cycles (vertical dashed lines). (a) The successive difference norm $\|\mathbf{z}^{\ell}_{\kcyc}-\mathbf{z}^{\ell}_{\kcyc-1}\|_2$ shows gating drives each layer's step size toward zero beyond the trained depth. (b) The normalized difference to each layer's deepest recorded state highlights the rate and timing of decay as key convergence diagnostics.}
    \label{fig:mech_state_convergence}
\end{figure}

\paragraph{Gates deliver the state-level stopping primitive that halting alone cannot.}

The per-layer step norms in Figures~\ref{fig:mech_state_convergence}a and~\ref{fig:mech_state_convergence_full}a reveal three distinct behaviors. Recurrence without GMR shows no consistent limiting behavior, although input injection keeps it bounded (Figure~\ref{app:fig_mech_injection}). Halt-only GMR changes only \emph{when} updates stop, not the underlying step size. GMR with residual gates provides what halt-only lacks: a state-level ``stop-moving'' mechanism that drives the step norm toward zero for the scalar and per-dim variants (\S\ref{app:sec_mech_analyses}).

\paragraph{Scalar and per-dimension residual gates deliver empirical fixed-point convergence.}

We use ``state convergence'' for variants whose update magnitude $\|\Delta\zlat\|_2$ approaches zero (Figures~\ref{fig:mech_state_convergence}a and~\ref{fig:mech_state_convergence_full}a) \emph{and} whose normalized residual $\|\mathbf{z}^{\ell}_{\kcyc}-\mathbf{z}^{\ell}_L\|_2/\|\mathbf{z}^{\ell}_L\|_2$ decays rapidly (Figures~\ref{fig:mech_state_convergence}b and~\ref{fig:mech_state_convergence_full}b). From this perspective, the four variants separate cleanly: scalar and per-dim gates reach near-zero residual within a few cycles of the trained horizon $\ncyc=6$, and their step sizes vanish. Halt-only decays only linearly with an unbounded norm while the rank-$16$ gate drifts radially. Cosine alignment to $\mathbf{z}^{\ell}_L$ is near-perfect for every variant and is therefore only an indirect diagnostic (Figure~\ref{fig:mech_state_convergence_full}c). Lastly, the induced attention geometry shows scalar and per-dim gates collapse cross-cycle attention distances past the trained horizon, whereas halt-only does not (Figure~\ref{app:fig_mech_post_halt_attn}). Empirical fixed-point convergence therefore requires a residual gate rather than the halting head alone.

\Needspace{18\baselineskip}
\subsubsection{Stages of Inference}
\label{sec:stages_of_inference}

\begin{wrapfigure}{r}{0.45\textwidth}
    \centering
    \vspace{-0.2in}
    \includegraphics[width=0.98\linewidth]{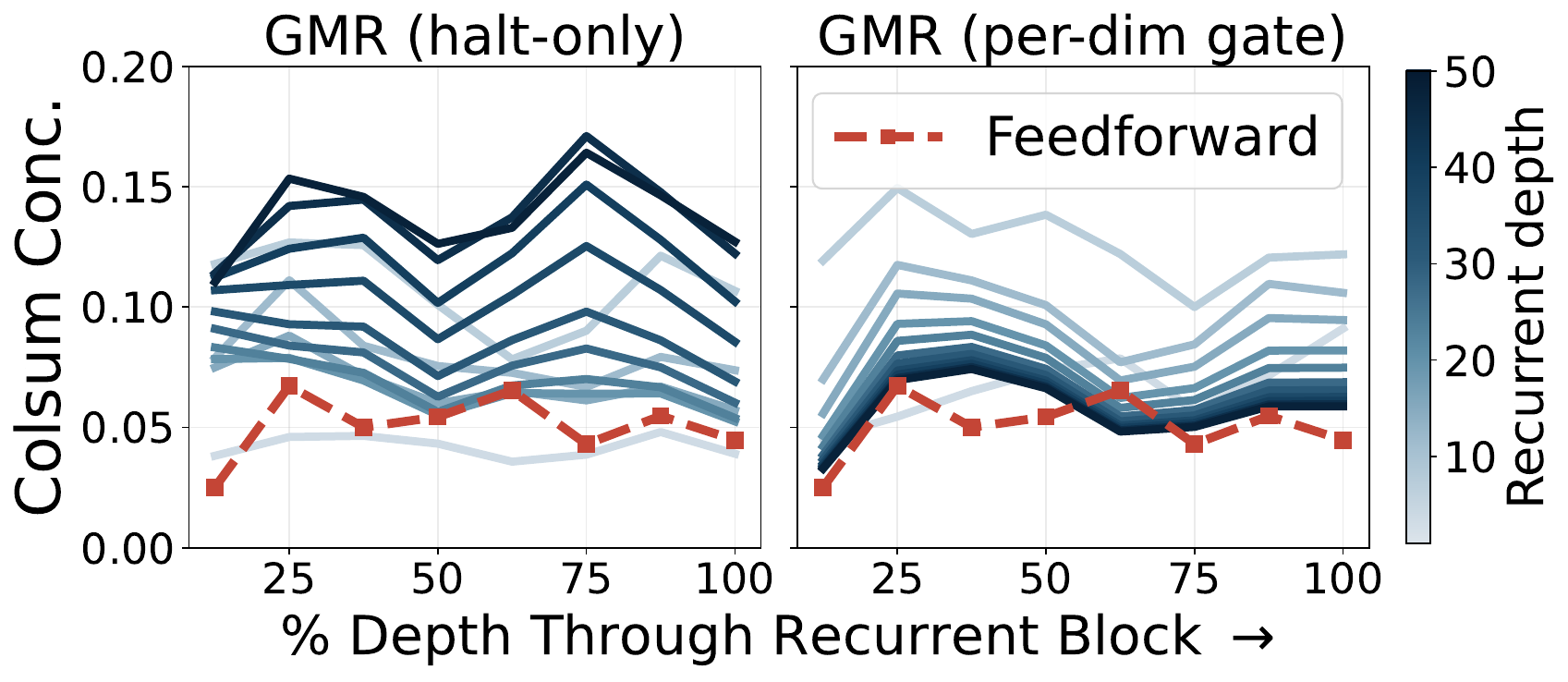}
    \caption{\textbf{ColSum concentration} through the recurrent block at $L_{\mathrm{eval}} = 50$. Cycles colored light (early) to dark (late). Dashed red line is the feedforward reference.}
    \label{fig:mech_ff_alignment}
    \vspace{-0.1in}
\end{wrapfigure}

Feedforward Transformer layers exhibit heterogeneous compute profiles, where early layers often cluster residues by local context, middle layers integrate over increasing spans, and late layers commit to global structural predictions~\citep{vig2021bertology,siji2026words}. We thus ask whether, for each recAMPLIFY variant, the tied block preserves this depth-heterogeneous mixing schedule within the trained recurrence depth $\ncyc=6$ and under extrapolation. Following \citet{queipodellano2026attention}, we measure attention geometry with ColSum concentration, $C \in [0, 1]$ (defined in \S\ref{app:colsum_definition}). A non-flat layerwise profile of $C$ indicates mixing heterogeneity across physical layers.

\paragraph{Within the trained depth, recursion preserves mixing heterogeneity.}

Within the trained horizon $k\leq6$ (solid curves in Figure~\ref{fig:mech_stages_of_inference}), all recAMPLIFY variants retain a structured,
non-flat ColSum profile across the eight tied core layers, with the earliest cycle closest to the feedforward reference.

\paragraph{Residual gates stabilize mixing beyond the trained horizon.}

Figure~\ref{fig:mech_stages_of_inference} shows that a stable per-layer state under deeper unrolling is marked by dashed extrapolated per-cycle profiles piled onto the solid trained range. Only the three GMR gates achieve this, all within a few cycles of the seen $L$. Stability alone, however, does not tell us whether the frozen profile reproduces the depth-heterogeneous mixing of a standard feedforward Transformer or has flattened into a distorted schedule. We next compare the two most contrastive variants at extrapolated depth: halt-only GMR, which never stabilizes, and per-dim GMR, which settles. Figure~\ref{fig:mech_ff_alignment} shows that only per-dim gating approaches the feedforward reference in both shape and amplitude. Lastly, the scalar and per-dimension gates stabilize attention geometry beyond the trained horizon, whereas halt-only and rank-16 do not (Figure~\ref{app:fig_mech_post_halt_attn}).

\afterpage{%
    \begin{figure}[t]
        \centering
        \begin{minipage}[c]{0.49\textwidth}
            \centering

            \includegraphics[
            width=\linewidth
            ]{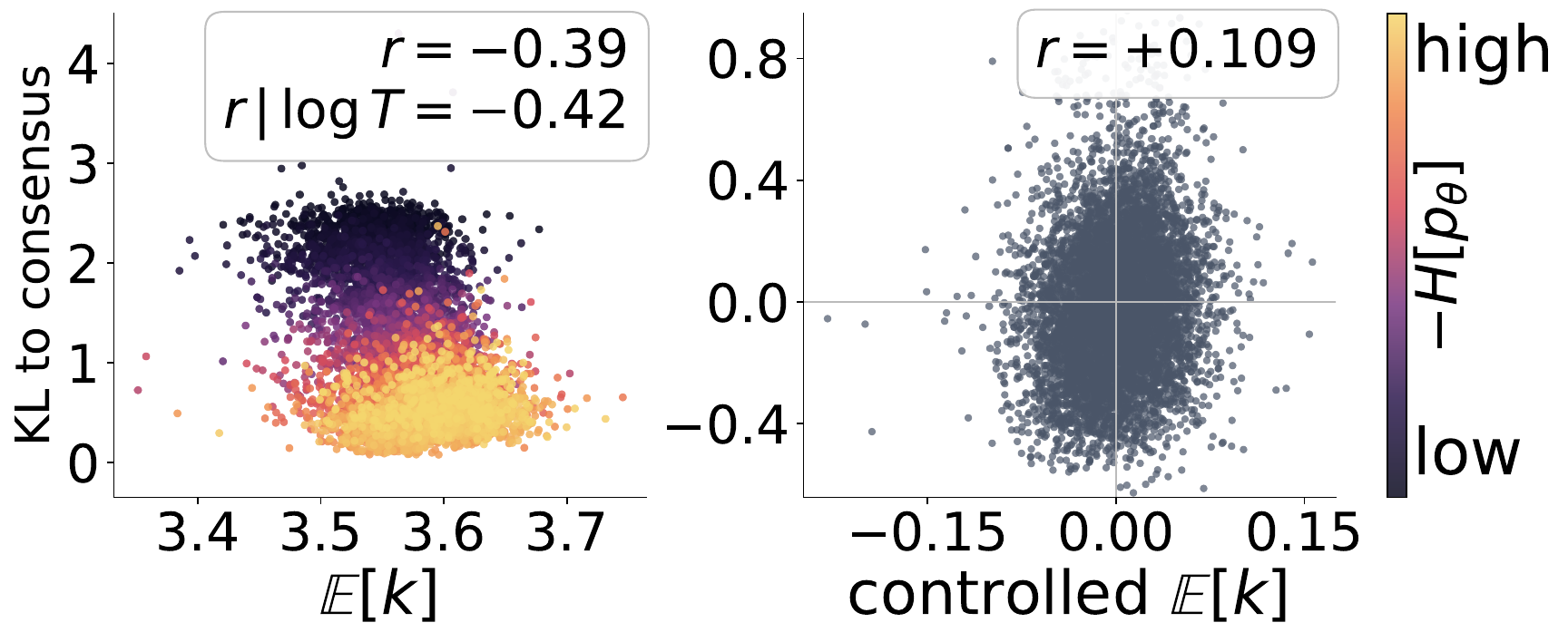}
            \caption{
               \textbf{Per-protein expected depth $\En$ vs mean per-position KL to MSA consensus (halt-only GMR, $6{,}868$ human proteins).} Colors code the model's prediction sharpness $-H[p_\theta]$, averaged over positions. Left: raw scatter; right: residuals of $\En$ and KL after regressing out $\log T$ and sharpness.
            }
            \label{fig:halt_sharpness_kl}

        \end{minipage}\hfill
        \begin{minipage}[c]{0.48\textwidth}
            \centering
            \captionof{table}{
                \textbf{Sequence composition Cohen's $d$ by controlled depth $\En$} on $10$K held-out UniProt sequences.}
            \label{tab:cross_task_property_signatures}

            \begin{adjustbox}{max width=\linewidth}
                \begin{tabular}{lc}
                    \toprule
                    Feature & Cohen's $d$ \\
                    \midrule
                    \multicolumn{2}{l}{\textbf{Per-protein composition ($+$ means over-pondered)}} \\
                    sequence Shannon entropy & $+1.47$ \\
                    hydrophobic (FILVMWY) fraction & $+1.45$ \\
                    aromatic (FWY) fraction & $+0.98$ \\
                    polar-neutral (STNQ) fraction & $-0.51$ \\
                    charged (DEKRH) fraction & $-0.28$ \\
                    histidine fraction & $+0.11$ \\
                    TM-like $19$-win / $100$ aa ($\text{KD} \geq 1.6$)
                    & $+0.35$ \\
                    sequence length & $-0.57$ \\
                    \midrule
                    \multicolumn{2}{l}{\textbf{Same-gene splice-isoform similarity}} \\
                    Isoforms of the same human gene ($335$ genes) & $+0.79$ \\
                    \bottomrule
                \end{tabular}
            \end{adjustbox}
        \end{minipage}
    \end{figure}
}

\subsubsection{Does the GMR Halt Depth Encode Biological Signal?}
\label{sec:halt_biology}

For per-protein halt depth $\En$ to be a reliable diagnostic signal, the model's distribution around an amino acid should align \emph{sharply} with what evolution prefers at the position. However, once both are controlled for sequence length $T$ and the model's own prediction sharpness $-H[p_\theta]$, the correlation between $\En$ and KL drops from $r = -0.39$ to $+0.11$, a reduction of $\approx72\%$ (Figure~\ref{fig:halt_sharpness_kl}, other tasks in Table~\ref{tab:halt_sharpness_multitask}). Halt depth therefore mostly reflects how confidently the model predicts a protein rather than its agreement with evolution per se. We thus define a \emph{controlled} $\En$, the residual of $\En$ after regressing out $\log T$ and $-H[p_\theta]$, and ask what it still encodes. Since both controls come from the model alone, no MSA is needed, so we compute the controlled $\En$ on $10$K held-out UniProt sequences, bucket them into top- and bottom-decile groups intersected across all four GMR variants, and report Cohen's $d$ between the buckets.

Table~\ref{tab:cross_task_property_signatures} identifies \emph{compositional diversity} as the most robust signal, with over-pondered proteins reflecting higher sequence entropy and hydrophobic content. Unusually hydrophobic composition is only viable inside a specific fold or membrane pocket, so recovering the wild-type residue there requires long-range structural evidence and thus more cycles. In contrast, extreme polar composition is an alphabet-reduction cue, where once a residue class dominates, most masked positions become predictable from local statistics and fewer cycles suffice. \S\ref{sec:biological_properties} shows hydrophobic content driving an overall depth increase while other class overrepresentations produce an inverted \emph{U-shape} consistent with alphabet reduction.

Next, we extend this test from per-protein composition to per-gene identity, and study how GMR depth varies across alternative splice-isoforms of the same human gene. Across $335$ human genes and $1{,}054$ isoforms, same-gene isoforms cluster nearly twice as tightly in halt depth as size-matched random pairs of proteins (Table~\ref{tab:cross_task_property_signatures}). This gap indicates that gates do not follow a fixed, input-independent decay schedule. Instead, the input governs the gate trajectory, at least at gene granularity. Moreover, what gates read is a property of the shared backbone rather than of exact residues since alternative isoforms differ only by one included or excluded functional domain, and $\En$ consistently tracks the compositional constraint of the retained portion across all family members (see \S\ref{app:isoform_data} for details).

\section{Conclusion}

We introduced recursive pretraining for test-time compute scaling in masked pLMs. With $56\%$ fewer unique parameters, the resulting recAMPLIFY improves CASP contact recovery, fold, localization and stability over AMPLIFY-$350$M while remaining close on DMS and non-target KL calibration. Our Gated Mixture Recurrence (GMR) unifies self-supervised halting and residual gating into one adaptive computation framework, reducing the expected per-protein cycle count by $32$--$40\%$ while improving contact recovery and most supervised probes over fixed-depth recursion. Mechanistically, scalar and per-dimension gates drive the recurrent state toward an empirical fixed point beyond the trained horizon, and only per-dimension gating recovers the feedforward layerwise mixing profile. Finally, after controlling for length and model confidence, halt depth tracks sequence entropy and hydrophobic composition, making it an interpretable compute allocator.

\subsection*{Acknowledgments and Disclosure of Funding}

We thank Lola Le Breton and David Heurtel-Depeiges for their valuable feedback on biology. We are grateful to Johan Obando-Ceron for his constructive feedback on mechanistic analyses of recAMPLIFY. \textbf{Funding.} Saurav Jha is supported by the IVADO postdoctoral fellowship and the Canada First Research Excellence Fund. Qianli Liao would like to thank Prof. Tomaso Poggio for helpful guidance, and was supported by MIT's K. Lisa Yang ICoN Center, Center for Brains, Minds and Machines (CBMM, NSF STC CCF-1231216), and McGovern Institute for Brain Research. Sarath Chandar is supported by the Canada CIFAR AI Chairs program, the Canada Research Chair in Lifelong Machine
Learning, and the NSERC Discovery Grant. Quentin Fournier is supported by Amgen. This research was enabled in part by compute resources provided by Amgen (\texttt{amgen.com}) and Mila (\texttt{mila.quebec}). \textbf{Competing interests.} Robert Vernon is an employee of Amgen. The other authors declare no competing interests.

\bibliography{iclr2027_conference}

@article{10.1093/bioinformatics/btm098,
	title        = {UniRef: comprehensive and non-redundant UniProt reference clusters},
	author       = {Suzek, Baris E. and Huang, Hongzhan and McGarvey, Peter and Mazumder, Raja and Wu, Cathy H.},
	year         = {2007},
	month        = {05},
	journal      = {Bioinformatics},
	volume       = {23},
	number       = {10},
	pages        = {1282--1288},
	doi          = {10.1093/bioinformatics/btm098},
	issn         = {1367-4803}
}

@article{act,
	title        = {Adaptive Computation Time for Recurrent Neural Networks},
	author       = {Graves, Alex},
	year         = {2016},
	journal      = {arXiv preprint arXiv:1603.08983},
	archiveprefix = {arXiv}
}

@article{amari1972learning,
	title        = {Learning Patterns and Pattern Sequences by Self-Organizing Nets of Threshold Elements},
	author       = {Amari, S.-I.},
	year         = {1972},
	journal      = {IEEE Transactions on Computers},
	volume       = {C-21},
	number       = {11},
	pages        = {1197--1206},
	doi          = {10.1109/T-C.1972.223477}
}

@article{amplify,
	title        = {Protein Language Models: Is Scaling Necessary?},
	author       = {Fournier, Quentin and Vernon, Robert M. and van der Sloot, Almer and Schulz, Benjamin and Chandar, Sarath and Langmead, Christopher James},
	year         = {2024},
	journal      = {bioRxiv},
	doi          = {10.1101/2024.09.23.614603}
}

@inproceedings{anil2022path,
	title        = {Path Independent Equilibrium Models Can Better Exploit Test-Time Computation},
	author       = {Anil, Cem and Pokle, Ashwini and Liang, Kaiqu and Treutlein, Johannes and Wu, Yuhuai and Bai, Shaojie and Kolter, J. Zico and Grosse, Roger B.},
	year         = {2022},
	booktitle    = {Advances in Neural Information Processing Systems},
	volume       = {35},
	pages        = {7796--7809}
}

@article{Armenteros2017deeploc,
	title        = {DeepLoc: prediction of protein subcellular localization using deep learning},
	author       = {Almagro Armenteros, José Juan and Sønderby, Casper Kaae and Sønderby, Søren Kaae and Nielsen, Henrik and Winther, Ole},
	year         = {2017},
	month        = {11},
	journal      = {Bioinformatics},
	volume       = {33},
	number       = {21},
	pages        = {3387--3395},
	doi          = {10.1093/bioinformatics/btx431},
	issn         = {1367-4803}
}

@article{baek2026generative,
	title        = {Generative recursive reasoning},
	author       = {Baek, Junyeob and Jo, Mingyu and Kim, Minsu and Ren, Mengye and Bengio, Yoshua and Ahn, Sungjin},
	year         = {2026},
	journal      = {arXiv preprint arXiv:2605.19376}
}

@article{bansal2022end,
	title        = {End-to-end algorithm synthesis with recurrent networks: Extrapolation without overthinking},
	author       = {Bansal, Arpit and Schwarzschild, Avi and Borgnia, Eitan and Emam, Zeyad and Huang, Furong and Goldblum, Micah and Goldstein, Tom},
	year         = {2022},
	journal      = {Advances in Neural Information Processing Systems},
	volume       = {35},
	pages        = {20232--20242}
}

@article{blayney2026mechanistic,
	title        = {A Mechanistic Analysis of Looped Reasoning Language Models},
	author       = {Blayney, Hugh and Arroyo, {\'A}lvaro and Obando-Ceron, Johan and Castro, Pablo Samuel and Courville, Aaron and Bronstein, Michael M. and Dong, Xiaowen},
	year         = {2026},
	journal      = {arXiv preprint arXiv:2604.11791},
	archiveprefix = {arXiv}
}

@article{chen2024xtrimopglm,
	title        = {xTrimoPGLM: unified 100-billion-parameter pretrained transformer for deciphering the language of proteins},
	author       = {Chen, Bo and Cheng, Xingyi and Li, Pan and Geng, Yangli-ao and Gong, Jing and Li, Shen and Bei, Zhilei and Tan, Xu and Wang, Boyan and Zeng, Xin and Liu, Chiming and Zeng, Aohan and Dong, Yuxiao and Tang, Jie and Song, Le},
	year         = {2025},
	journal      = {Nature Methods},
	volume       = {22},
	number       = {5},
	pages        = {1028--1039},
	doi          = {10.1038/s41592-025-02636-z},
	isbn         = {1548-7105},
	date         = {2025/05/01},
	id           = {Chen2025}
}

@article{cheng2024training,
	title        = {Training compute-optimal protein language models},
	author       = {Cheng, Xingyi and Chen, Bo and Li, Pan and Gong, Jing and Tang, Jie and Song, Le},
	year         = {2024},
	journal      = {Advances in Neural Information Processing Systems},
	volume       = {37},
	pages        = {69386--69418}
}

@article{cobbe2021gsm8k,
	title        = {Training Verifiers to Solve Math Word Problems},
	author       = {Cobbe, Karl and Kosaraju, Vineet and Bavarian, Mohammad and Chen, Mark and Jun, Heewoo and Kaiser, Lukasz and Plappert, Matthias and Tworek, Jerry and Hilton, Jacob and Nakano, Reiichiro and Hesse, Christopher and Schulman, John},
	year         = {2021},
	journal      = {arXiv preprint arXiv:2110.14168}
}

@article{cornelio2025need,
	title        = {The need for verification in artificial intelligence-driven scientific discovery},
	author       = {Cornelio, Cristina and Ito, Takuya and Cory-Wright, Ryan and Dash, Sanjeeb and Horesh, Lior},
	year         = {2026},
	month        = {04},
	journal      = {Philosophical Transactions of the Royal Society A: Mathematical, Physical and Engineering Sciences},
	volume       = {384},
	number       = {2317},
	pages        = {20240591},
	doi          = {10.1098/rsta.2024.0591},
	issn         = {1364-503X}
}

@inproceedings{dallago2021flip,
	title        = {{FLIP}: Benchmark tasks in fitness landscape inference for proteins},
	author       = {Christian Dallago and Jody Mou and Kadina E Johnston and Bruce Wittmann and Nick Bhattacharya and Samuel Goldman and Ali Madani and Kevin K Yang},
	year         = {2021},
	booktitle    = {Thirty-fifth Conference on Neural Information Processing Systems Datasets and Benchmarks Track (Round 2)}
}

@inproceedings{devlin2019bert,
	title        = {Bert: Pre-training of deep bidirectional transformers for language understanding},
	author       = {Devlin, Jacob and Chang, Ming-Wei and Lee, Kenton and Toutanova, Kristina},
	year         = {2019},
	booktitle    = {Proceedings of the 2019 conference of the North American chapter of the association for computational linguistics: human language technologies, volume 1 (long and short papers)},
	pages        = {4171--4186}
}

@inproceedings{didi2026flip2,
	title        = {{FLIP}2: Expanding Protein Fitness Landscape Benchmarks for Real-World Machine Learning Applications},
	author       = {Kieran Didi and Sarah Alamdari and Alex Xijie Lu and Bruce James Wittmann and Kadina E Johnston and Ava P Amini and Ali Madani and Maya Czeneszew and Christian Dallago and Kevin K Yang},
	year         = {2026},
	booktitle    = {Forty-third International Conference on Machine Learning}
}

@article{elnaggar2021prottrans,
	title        = {{ProtTrans}: Toward Understanding the Language of Life Through Self-Supervised Learning},
	author       = {Elnaggar, Ahmed and Heinzinger, Michael and Dallago, Christian and Rehawi, Ghalia and Wang, Yu and Jones, Llion and Gibbs, Tom and Feher, Tamas and Angerer, Christoph and Steinegger, Martin and Bhowmik, Debsindhu and Rost, Burkhard},
	year         = {2022},
	journal      = {IEEE Transactions on Pattern Analysis and Machine Intelligence},
	volume       = {44},
	number       = {10},
	pages        = {7112--7127},
	doi          = {10.1109/TPAMI.2021.3095381}
}

@article{esm2,
	title        = {Evolutionary-Scale Prediction of Atomic-Level Protein Structure with a Language Model},
	author       = {Lin, Zeming and Akin, Halil and Rao, Roshan and Hie, Brian and Zhu, Zhongkai and Lu, Wenting and Smetanin, Nikita and Verkuil, Robert and Kabeli, Ori and Shmueli, Yaniv and dos Santos Costa, Allan and Fazel-Zarandi, Maryam and Sercu, Tom and Candido, Salvatore and Rives, Alexander},
	year         = {2023},
	journal      = {Science},
	volume       = {379},
	number       = {6637},
	pages        = {1123--1130},
	doi          = {10.1126/science.ade2574}
}

@article{esmc,
	title        = {Language Modeling Materializes a World Model of Protein Biology},
	author       = {Candido, Salvatore and Hayes, Thomas and Derry, Alexander and Rao, Roshan and Lin, Zeming and Verkuil, Robert and Wu, Bryan Z. and Lee, Jin Sub and Bruguera, Elise S. and Keval, Jehan A. and Kopylov, Mykhailo and Pak, John E. and Wu, Wesley and Thomas, Neil and Mataraso, Samson and Hsu, Alvin and Trotman-Grant, Ashton C. and Fatras, Kilian and dos Santos Costa, Allan and Badkundri, Rohil and Ak{\i}n, Halil and Oktay, Deniz and Deaton, Jonathan and Montabana, Elizabeth and Sitwala, Hrishita and Yu, Yue and Wiggert, Marius and Carlin, Dylan Alexander and Goering, Anthony W. and Blazejewski, Tomasz and Sandora, McCullen and Hla, Michael and Jia, Tina Z. and Kloker, Leon H. and Sofroniew, Nicholas J. and Uehara, Masatoshi and Pannu, Jassi and Bachas, Sharrol and Liu, Daniel S. and Sercu, Tom and Rives, Alexander},
	year         = {2026},
	journal      = {bioRxiv},
	publisher    = {Cold Spring Harbor Laboratory},
	doi          = {10.64898/2026.06.03.729735},
	elocation-id = {2026.06.03.729735}
}

@article{fein2026solve,
	title        = {Solve the Loop: Attractor Models for Language and Reasoning},
	author       = {Fein-Ashley, Jacob and Rashidinejad, Paria},
	year         = {2026},
	journal      = {arXiv preprint arXiv:2605.12466}
}

@article{ferruz2022protgpt2,
	title        = {{ProtGPT2} Is a Deep Unsupervised Language Model for Protein Design},
	author       = {Ferruz, Noelia and Schmidt, Steffen and H{\"o}cker, Birte},
	year         = {2022},
	journal      = {Nature Communications},
	volume       = {13},
	number       = {1},
	pages        = {4348},
	doi          = {10.1038/s41467-022-32007-7}
}

@article{Fox2013SCOPeSC,
	title        = {SCOPe: Structural Classification of Proteins—extended, integrating SCOP and ASTRAL data and classification of new structures},
	author       = {Fox, Naomi K. and Brenner, Steven E. and Chandonia, John-Marc},
	year         = {2014},
	month        = {01},
	journal      = {Nucleic Acids Research},
	volume       = {42},
	number       = {D1},
	pages        = {D304-D309},
	doi          = {10.1093/nar/gkt1240},
	issn         = {0305-1048}
}

@inproceedings{geiping2026scaling,
	title        = {Scaling up Test-Time Compute with Latent Reasoning: A Recurrent Depth Approach},
	author       = {Geiping, Jonas and McLeish, Sean and Jain, Neel and Kirchenbauer, John and Singh, Siddharth and Bartoldson, Brian and Kailkhura, Bhavya and Bhatele, Abhinav and Goldstein, Tom},
	year         = {2025},
	booktitle    = {Advances in Neural Information Processing Systems},
	publisher    = {Curran Associates, Inc.},
	volume       = {38, Main Conference},
	pages        = {41340--41391},
	doi          = {10.52202/085713-1380},
	editor       = {D. Belgrave and C. Zhang and H. Lin and R. Pascanu and P. Koniusz and M. Ghassemi and N. Chen}
}

@inproceedings{hao2025training,
	title        = {Training Large Language Models to Reason in a Continuous Latent Space},
	author       = {Shibo Hao and Sainbayar Sukhbaatar and DiJia Su and Xian Li and Zhiting Hu and Jason E Weston and Yuandong Tian},
	year         = {2025},
	booktitle    = {Second Conference on Language Modeling}
}

@article{hou2026understanding,
	title        = {Understanding language model scaling for protein fitness prediction},
	author       = {Hou, Chao and Liu, Di and Zafar, Aziz and Shen, Yufeng},
	year         = {2026},
	journal      = {Nature Computational Science},
	volume       = {6},
	number       = {7},
	pages        = {778--788},
	doi          = {10.1038/s43588-026-01010-z},
	isbn         = {2662-8457},
	date         = {2026/07/01},
	id           = {Hou2026}
}

@article{hrm,
	title        = {Hierarchical Reasoning Model},
	author       = {Wang, Guan and Li, Jin and Sun, Yuhao and Chen, Xing and Liu, Changling and Wu, Yue and Lu, Meng and Song, Sen and Abbasi-Yadkori, Yasin},
	year         = {2025},
	journal      = {arXiv preprint arXiv:2506.21734},
	archiveprefix = {arXiv}
}

@article{jumper2021alphafold,
	title        = {Highly accurate protein structure prediction with AlphaFold},
	author       = {Jumper, John and Evans, Richard and Pritzel, Alexander and Green, Tim and Figurnov, Michael and Ronneberger, Olaf and Tunyasuvunakool, Kathryn and Bates, Russ and {\v{Z}}{\'\i}dek, Augustin and Potapenko, Anna and others},
	year         = {2021},
	journal      = {nature},
	publisher    = {Nature Publishing Group UK London},
	volume       = {596},
	number       = {7873},
	pages        = {583--589}
}

@article{Klausen2018NetSurfP20IP,
	title        = {NetSurfP-2.0: Improved prediction of protein structural features by integrated deep learning},
	author       = {Klausen, Michael Schantz and Jespersen, Martin Closter and Nielsen, Henrik and Jensen, Kamilla Kjærgaard and Jurtz, Vanessa Isabell and Sønderby, Casper Kaae and Sommer, Morten Otto Alexander and Winther, Ole and Nielsen, Morten and Petersen, Bent and Marcatili, Paolo},
	year         = {2019},
	journal      = {Proteins: Structure, Function, and Bioinformatics},
	volume       = {87},
	number       = {6},
	pages        = {520--527},
	doi          = {https://doi.org/10.1002/prot.25674}
}

@inproceedings{kohli2026loop,
	title        = {Loop, Think, \& Generalize: Implicit Reasoning in Recurrent-Depth Transformers},
	author       = {Harsh Kohli and Srinivasan Parthasarathy and Huan Sun and Yuekun Yao},
	year         = {2026},
	booktitle    = {Third Conference on Language Modeling}
}

@article{LeBreton2026,
	title        = {pLM representations unlock metagenomic space beyond homology},
	author       = {Le Breton, Lola and Heurtel-Depeiges, David and Millar, Douglas C and Zetzsche, Lara E and Vernon, Robert M and Langmead, Christopher James and Chandar, Sarath and Fournier, Quentin},
	year         = {2026},
	journal      = {bioRxiv},
	publisher    = {Cold Spring Harbor Laboratory},
	doi          = {10.64898/2026.07.28.739874},
	elocation-id = {2026.07.28.739874}
}

@inproceedings{looped,
	title        = {Looped Transformers as Programmable Computers},
	author       = {Giannou, Angeliki and Rajput, Shashank and Sohn, Jy-Yong and Lee, Kangwook and Lee, Jason D. and Papailiopoulos, Dimitris},
	year         = {2023},
	booktitle    = {Proceedings of the 40th International Conference on Machine Learning},
	publisher    = {PMLR},
	volume       = {202},
	pages        = {11398--11442}
}

@inproceedings{loshchilov2018decoupled,
	title        = {Decoupled Weight Decay Regularization},
	author       = {Ilya Loshchilov and Frank Hutter},
	year         = {2019},
	booktitle    = {International Conference on Learning Representations}
}

@article{lupo2023categoricaljacobian,
	title        = {Protein language models learn evolutionary statistics of interacting sequence motifs},
	author       = {Zhidian Zhang  and Hannah K. Wayment-Steele  and Garyk Brixi  and Haobo Wang  and Dorothee Kern  and Sergey Ovchinnikov},
	year         = {2024},
	journal      = {Proceedings of the National Academy of Sciences},
	volume       = {121},
	number       = {45},
	pages        = {e2406285121},
	doi          = {10.1073/pnas.2406285121}
}

@inproceedings{meier2021esm1v,
	title        = {Language Models Enable Zero-Shot Prediction of the Effects of Mutations on Protein Function},
	author       = {Meier, Joshua and Rao, Roshan and Verkuil, Robert and Liu, Jason and Sercu, Tom and Rives, Alexander},
	year         = {2021},
	booktitle    = {Advances in Neural Information Processing Systems},
	volume       = {34},
	pages        = {29287--29303}
}

@article{murzin1995scop,
	title        = {SCOP: a structural classification of proteins database for the investigation of sequences and structures},
	author       = {Murzin, Alexey G and Brenner, Steven E and Hubbard, Tim and Chothia, Cyrus},
	year         = {1995},
	journal      = {Journal of molecular biology},
	publisher    = {Elsevier},
	volume       = {247},
	number       = {4},
	pages        = {536--540}
}

@article{nakano1972associatron,
	title        = {Associatron-A Model of Associative Memory},
	author       = {Nakano, Kaoru},
	year         = {1972},
	journal      = {IEEE Transactions on Systems, Man, and Cybernetics},
	volume       = {SMC-2},
	number       = {3},
	pages        = {380--388},
	doi          = {10.1109/TSMC.1972.4309133}
}

@inproceedings{NEURIPS2021_3501672e,
	title        = {Can You Learn an Algorithm?  Generalizing from Easy to Hard Problems with Recurrent Networks},
	author       = {Schwarzschild, Avi and Borgnia, Eitan and Gupta, Arjun and Huang, Furong and Vishkin, Uzi and Goldblum, Micah and Goldstein, Tom},
	year         = {2021},
	booktitle    = {Advances in Neural Information Processing Systems},
	publisher    = {Curran Associates, Inc.},
	volume       = {34},
	pages        = {6695--6706},
	editor       = {M. Ranzato and A. Beygelzimer and Y. Dauphin and P.S. Liang and J. Wortman Vaughan}
}

@article{nijkamp2023progen2,
	title        = {{ProGen2}: Exploring the Boundaries of Protein Language Models},
	author       = {Nijkamp, Erik and Ruffolo, Jeffrey A. and Weinstein, Eli N. and Naik, Nikhil and Madani, Ali},
	year         = {2023},
	journal      = {Cell Systems},
	volume       = {14},
	number       = {11},
	pages        = {968--978.e3},
	doi          = {10.1016/j.cels.2023.10.002}
}

@article{olsen2022observed,
	title        = {Observed Antibody Space: A diverse database of cleaned, annotated, and translated unpaired and paired antibody sequences},
	author       = {Olsen, Tobias H and Boyles, Fergus and Deane, Charlotte M},
	year         = {2022},
	journal      = {Protein Science},
	publisher    = {Wiley Online Library},
	volume       = {31},
	number       = {1},
	pages        = {141--146}
}

@inproceedings{pondernet,
	title        = {{PonderNet}: Learning to Ponder},
	author       = {Banino, Andrea and Balaguer, Jan and Blundell, Charles},
	year         = {2021},
	booktitle    = {ICML 2021 Workshop on Automated Machine Learning},
	archiveprefix = {arXiv}
}

@article{prairie2026parcae,
	title        = {Parcae: Scaling laws for stable looped language models},
	author       = {Prairie, Hayden and Novack, Zachary and Berg-Kirkpatrick, Taylor and Fu, Daniel Y},
	year         = {2026},
	journal      = {arXiv preprint arXiv:2604.12946}
}

@inproceedings{proteingym,
	title        = {ProteinGym: Large-Scale Benchmarks for Protein Fitness Prediction and Design},
	author       = {Notin, Pascal and Kollasch, Aaron and Ritter, Daniel and van Niekerk, Lood and Paul, Steffanie and Spinner, Han and Rollins, Nathan and Shaw, Ada and Orenbuch, Rose and Weitzman, Ruben and Frazer, Jonathan and Dias, Mafalda and Franceschi, Dinko and Gal, Yarin and Marks, Debora},
	year         = {2023},
	booktitle    = {Advances in Neural Information Processing Systems},
	publisher    = {Curran Associates, Inc.},
	volume       = {36},
	pages        = {64331--64379},
	editor       = {A. Oh and T. Neumann and A. Globerson and K. Saenko and M. Hardt and S. Levine}
}

@inproceedings{queipodellano2026attention,
	title        = {Attention Sinks and Compression Valleys in {LLM}s are Two Sides of the Same Coin},
	author       = {Enrique Queipo-de-Llano and Alvaro Arroyo and Federico Barbero and Xiaowen Dong and Michael M. Bronstein and Yann LeCun and Ravid Shwartz-Ziv},
	year         = {2026},
	booktitle    = {The Fourteenth International Conference on Learning Representations}
}

@article{rao2019evaluating,
	title        = {Evaluating protein transfer learning with TAPE},
	author       = {Rao, Roshan and Bhattacharya, Nicholas and Thomas, Neil and Duan, Yan and Chen, Peter and Canny, John and Abbeel, Pieter and Song, Yun},
	year         = {2019},
	journal      = {Advances in neural information processing systems},
	volume       = {32}
}

@article{rives2021biological,
	title        = {Biological Structure and Function Emerge from Scaling Unsupervised Learning to 250 Million Protein Sequences},
	author       = {Rives, Alexander and Meier, Joshua and Sercu, Tom and Goyal, Siddharth and Lin, Zeming and Liu, Jason and Guo, Demi and Ott, Myle and Zitnick, C. Lawrence and Ma, Jerry and Fergus, Rob},
	year         = {2021},
	journal      = {Proceedings of the National Academy of Sciences},
	volume       = {118},
	number       = {15},
	pages        = {e2016239118},
	doi          = {10.1073/pnas.2016239118}
}

@article{rocklin2017massivelyParallel,
	title        = {Global analysis of protein folding using massively parallel design, synthesis, and testing},
	author       = {Gabriel J. Rocklin  and Tamuka M. Chidyausiku  and Inna Goreshnik  and Alex Ford  and Scott Houliston  and Alexander Lemak  and Lauren Carter  and Rashmi Ravichandran  and Vikram K. Mulligan  and Aaron Chevalier  and Cheryl H. Arrowsmith  and David Baker},
	year         = {2017},
	journal      = {Science},
	volume       = {357},
	number       = {6347},
	pages        = {168--175},
	doi          = {10.1126/science.aan0693}
}

@inproceedings{siji2026words,
	title        = {From Words to Amino Acids: Does the Curse of Depth Persist?},
	author       = {Aleena Siji and Amir Mohammad Karimi Mamaghan and Ferdinand Kapl and Tobias H{\"o}ppe and Emmanouil Angelis and Andrea Dittadi and Maurice Brenner and Michael Heinzinger and Karl Henrik Johansson and Kaitlin Maile and Johannes von Oswald and Stefan Bauer},
	year         = {2026},
	booktitle    = {Learning Meaningful Representations of Life (LMRL) Workshop at ICLR 2026}
}

@unknown{srm,
	title        = {Simple Recursive Model: Simplified, Single-State Reasoning with Skip Connections},
	author       = {Liao, Qianli and Poggio, Tomaso},
	year         = {2026},
	month        = {01},
	doi          = {10.21203/rs.3.rs-8492126/v1}
}

@article{su2024roformer,
	title        = {Roformer: Enhanced transformer with rotary position embedding},
	author       = {Su, Jianlin and Ahmed, Murtadha and Lu, Yu and Pan, Shengfeng and Bo, Wen and Liu, Yunfeng},
	year         = {2024},
	journal      = {Neurocomputing},
	publisher    = {Elsevier},
	volume       = {568},
	pages        = {127063}
}

@article{trm,
	title        = {Less Is More: Recursive Reasoning with Tiny Networks},
	author       = {Jolicoeur-Martineau, Alexia},
	year         = {2025},
	journal      = {arXiv preprint arXiv:2510.04871},
	archiveprefix = {arXiv}
}

@inproceedings{ut,
	title        = {Universal Transformers},
	author       = {Dehghani, Mostafa and Gouws, Stephan and Vinyals, Oriol and Uszkoreit, Jakob and Kaiser, Lukasz},
	year         = {2019},
	booktitle    = {International Conference on Learning Representations}
}

@article{vaswani2017attention,
	title        = {Attention is all you need},
	author       = {Vaswani, Ashish and Shazeer, Noam and Parmar, Niki and Uszkoreit, Jakob and Jones, Llion and Gomez, Aidan N and Kaiser, Lukasz and Polosukhin, Illia},
	year         = {2017},
	journal      = {Advances in neural information processing systems},
	volume       = {30}
}

@inproceedings{vig2021bertology,
	title        = {{BERT}ology Meets Biology: Interpreting Attention in Protein Language Models},
	author       = {Jesse Vig and Ali Madani and Lav R. Varshney and Caiming Xiong and Richard Socher and Nazneen Rajani},
	year         = {2021},
	booktitle    = {International Conference on Learning Representations}
}

@inproceedings{wang2023selfconsistency,
	title        = {Self-Consistency Improves Chain of Thought Reasoning in Language Models},
	author       = {Xuezhi Wang and Jason Wei and Dale Schuurmans and Quoc V Le and Ed H. Chi and Sharan Narang and Aakanksha Chowdhery and Denny Zhou},
	year         = {2023},
	booktitle    = {The Eleventh International Conference on Learning Representations}
}

@inproceedings{wei2022cot,
	title        = {Chain-of-Thought Prompting Elicits Reasoning in Large Language Models},
	author       = {Wei, Jason and Wang, Xuezhi and Schuurmans, Dale and Bosma, Maarten and Ichter, Brian and Xia, Fei and Chi, Ed and Le, Quoc V. and Zhou, Denny},
	year         = {2022},
	booktitle    = {Advances in Neural Information Processing Systems},
	volume       = {35},
	pages        = {24824--24837}
}

@inproceedings{yoran2023answering,
	title        = {Answering questions by meta-reasoning over multiple chains of thought},
	author       = {Yoran, Ori and Wolfson, Tomer and Bogin, Ben and Katz, Uri and Deutch, Daniel and Berant, Jonathan},
	year         = {2023},
	booktitle    = {Proceedings of the 2023 Conference on Empirical Methods in Natural Language Processing},
	pages        = {5942--5966}
}

@article{Cheng2023CoevolutionbasedPO,
  title={Co-evolution-based prediction of metal-binding sites in proteomes by machine learning},
  author={Yaozhong Cheng and Hao-Bo Wang and Hua Xu and Yuan Liu and Bin-bin Ma and Xuemin Chen and Xin Zeng and Xianghe Wang and Bo Wang and Carina Shiau and Sergey Ovchinnikov and Xiao-Dong Su and Chu Wang},
  journal={Nature Chemical Biology},
  year={2023},
  volume={19},
  pages={548-555},
  url={https://api.semanticscholar.org/CorpusID:255469189}
}

@article{yuan2026casp16,
	title        = {CASP16 protein monomer structure prediction assessment},
	author       = {Yuan, Rongqing and Zhang, Jing and Kryshtafovych, Andriy and Schaeffer, R Dustin and Zhou, Jian and Cong, Qian and Grishin, Nick V},
	year         = {2026},
	journal      = {Proteins: Structure, Function, and Bioinformatics},
	publisher    = {Wiley Online Library},
	volume       = {94},
	number       = {1},
	pages        = {86--105}
}

@article{liao2016bridging,
  title={Bridging the gaps between residual learning, recurrent neural networks and visual cortex},
  author={Liao, Qianli and Poggio, Tomaso},
  journal={arXiv preprint arXiv:1604.03640},
  year={2016}
}

@article{chandonia2022scope,
	title        = {{SCOPe}: improvements to the structural classification of proteins - extended database to facilitate variant interpretation and machine learning},
	author       = {Chandonia, John-Marc and Guan, Lindsey and Lin, Shiangyi and Yu, Changhua and Fox, Naomi K and Brenner, Steven E},
	year         = 2022,
	journal      = {Nucleic Acids Research},
	volume       = 50,
	number       = {D1},
	pages        = {D553-D559},
	doi          = {10.1093/nar/gkab1054},
	issn         = {1362-4962},
	shortjournal = {Nucleic Acids Res},
	pmid         = 34850923
}
\bibliographystyle{plainnat}
\clearpage
\appendix

\section{Forward-Pass Pseudocode for recAMPLIFY With GMR}
\label{app:pseudocode}

Listing~\ref{alg:ar} shows a single training-time forward pass of recAMPLIFY for all four GMR variants (halt-only and the three residual gate resolutions). Notation follows \S\ref{sec:gmr}: \texttt{sg} is the stopgrad operator, \texttt{halt\_head} and \texttt{gate\_head} are linear heads that produce $\lam$ and $g^{\text{ch}}$, \texttt{Norm} is RMSNorm, and $\operatorname{Map\_A}$ is the channel-to-feature broadcast for $c = 1$ scalar gate, row-stochastic low-rank projection, or the identity for per-dim gate. The final \texttt{Loss} is the MLM cross-entropy on the mixture-decoded state $\bar{\zlat}$ minus the halt-distribution entropy regularizer $H$ from~equation~\ref{eq:ponder}. The backward pass uses standard autograd with stopgrad boundaries as marked.

\begin{lstlisting}[
    float=htbp,
    style=pseudocode,
    caption={\textbf{Python-styled pseudocode} for a single training-time forward pass of recAMPLIFY with Gated Mixture Recurrence (GMR). $\operatorname{Map\_A}$ broadcasts a scalar gate, applies the learned low-rank assignment for the low-rank gate, or is the identity for per-dimension gate.},
    label={alg:ar}
]
x, z = Prelude(Embed(u)), z_init
inj  = x + f_c(z_init)                         # f = optional SRM-styled context encoder f_c

if mode == "halt_only_GMR":                            # scalar, per-sequence halting
    p_left, z_bar, p = 1, 0, []
    for k in 1..L:
        z = f(z + inj)
        lam_k = sigmoid(halt_head(sg(z_bos)))
        if k == L: lam_k = 1                   # absorb remaining mass
        p_k = p_left * lam_k
        p_left *= 1 - lam_k
        p.append(p_k)
        z_bar += p_k * z
        if all_sequences(p_left < eps): break
else:                # scalar | low_rank | per_dim gate
    S_c, z_bar, p_ch = ones(C), 0, []
    for k in 1..L:
        z_new = f(z + inj)                   # recurrent proposal
        g_ch = sigmoid(gate_head(sg(z_new_bos)))
        if k == L: g_ch = 0                   # absorb remaining channel mass
        S_D = Map_A(S_c)                     # map accumulated survival to features
        z = z + S_D * (z_new - z)
        p_k_ch = S_c * (1 - g_ch)              # channel-wise halt mass
        p_ch.append(p_k_ch)
        w_k = Map_A(p_k_ch) if per_dim_decode else mean(p_k_ch)
        z_bar += w_k * Norm(z)
        S_c *= g_ch
        if all_sequences(max_channel(S_c) < eps): break

Loss  = L_MLM(LM_head(Norm(Coda(z_bar))), labels)
Loss -= entropy(p if mode == "halt_only_GMR" else p_ch)
return Loss
\end{lstlisting}

\section{Pretraining Setup}
\label{app:pretraining_setup}

Following AMPLIFY~\citep{amplify}, we train for $100$K steps with a $210$B-token budget on $8$ NVIDIA H$100$ GPUs. This includes $32{,}768$ tokens per device, packed into a single \textsc{FlashAttention} varlen buffer, with $8$ gradient-accumulation steps on the training mixture of UniRef100~\citep{10.1093/bioinformatics/btm098} for the general proteome, SCOP~\citep{chandonia2022scope} for structural domains, and OAS~\citep{olsen2022observed} for antibody sequences. We use the standard MLM objective with a masking rate of $0.15$ and truncate input sequences to a maximum length of $512$. Optimization uses AdamW~\citep{loshchilov2018decoupled} with ($\beta_1 = 0.9$, $\beta_2 = 0.95$) and a weight decay factor of $10^{-2}$. We clip gradients to a maximum norm of $1.0$ and use a cosine learning-rate schedule with a $10\%$ minimum-LR floor and a $1{,}000$-step warm-up. All models, including the AMPLIFY baselines, use a peak learning rate of $10^{-3}$. Architecturally, in the $1$M-step training setup, our $155$M-parameter backbone matches AMPLIFY-$350$M's hidden width of $960$ with $15$ attention heads, each of dimension $64$, alongside an FFN width of $2{,}560$ and RoPE~\citep{su2024roformer} with $\theta = 10{,}000$. Following \citet{pondernet}, we set $\varepsilon = 0.001$.

\paragraph{Dataset.}

We pretrain on a mixture of three corpora: UniRef$100$~\citep{10.1093/bioinformatics/btm098} for the general proteome, SCOP~\citep{chandonia2022scope} for structural domain diversity, and OAS~\citep{olsen2022observed} for antibody sequences. During training, we do not apply explicit resampling or per-corpus weighting. Each sequence therefore appears in the training stream in proportion to the raw size of its source corpus, so UniRef$100$ contributes the vast majority of samples, while SCOP and OAS increase structural and antibody coverage. Training batches are formed by a token-budget sampler that greedily packs variable-length sequences into blocks of $32{,}768$ tokens per device, with per-epoch shuffling over the concatenated index, so every batch is a well-mixed sample drawn across all three corpora. The three held-out validation splits are evaluated separately, yielding one perplexity (PPL) curve per corpus (Figure~\ref{app:ppl_devset}).

\paragraph{Why do we choose the pretraining recipe of AMPLIFY over ESM-2's UniRef50/90?}
\label{app:pretraining_data_choice}

\citet{amplify} argue that UniRef50, obtained by clustering UniRef100 at $50\%$ identity, systematically distorts the natural sequence distribution in three ways. First, cluster size in UniRef100 is an observational signal. Since large clusters correspond to families for which many related sequences have been deposited, their redundancy reflects real evidence for the underlying subspace of the fitness landscape. However, $50\%$-identity clustering discards this signal and upweights singletons, which are more likely to arise from sequencing errors. Second, the fraction of ambiguous reads, fragmentary sequences, and hypothetical proteins in UniRef50 is markedly higher than in UniRef100. Third, specific functional classes are depleted under $50\%$-identity clustering. AMPLIFY's own $350$M model, trained on a mixture of UniRef100, OAS, and SCOP, either matches or exceeds ESM-2~$15$B on the proteome-completeness-weighted validation despite the $ \sim 42 \times $ scale gap, which we take as empirical confirmation of the recipe.

\section{Evaluation Setup}
\label{app:eval_details}

\subsection{Evaluation Tasks}

We describe the four tasks introduced in \S\ref{sec:experiments} in detail. The full pipeline is a two-stage protocol: a zero-shot stage in which the model is queried via its frozen MLM head, and a supervised stage in which frozen per-token embeddings are extracted once and used to train a downstream MLP head with a controlled hyperparameter search.

\subsubsection{Zero-Shot Tasks}
\label{app:zero_shot}

We evaluate our frozen MLM heads on the following three zero-shot tasks:

\paragraph{KL-calibration to human MSA consensus (T, NT)~\citep{LeBreton2026}.}

We consider $6{,}868$ human proteins. For each position $i$ with an MSA consensus profile, we mask that position and read the model’s per-position $20$-way amino-acid distribution $p_{\theta}(\cdot \mid \seq_{\setminus i})$ from the MLM head. Independently, we compute a consensus profile $\mathbf{f}_i \in \Delta^{19}$ at each position from a curated multiple sequence alignment (MSA) of homologs. For each position, we report the scaled calibration score: $\mathrm{cal}_i = 100\,\exp \big(-\KL( \mathbf{f}_i \| p_{\theta})\big) \in [0, 100],$ where higher is better. We separately average across positions where the wild-type residue matches the consensus (target, T) and where it does not (non-target, NT). This split isolates the target (T) predictions that the model is expected to recover accurately from the non-target (NT) predictions that require distinguishing evolutionarily tolerated variants. Higher T rewards accurate consensus recovery, while higher NT implies better generalization.

\paragraph{ProteinGym DMS mutation-effect ranking.}

We use the full substitution benchmark from ProteinGym~\citep{proteingym}, which contains $217$ deep mutational scanning assays covering $2{,}465{,}767$ single- and multi-mutant variants. For each assay, mutation effects are scored using the ESM-2-style masked-marginal formula popularized by \citet{meier2021esm1v}:
\begin{equation}
    s(\mathrm{mt}) = \sum_{i \in \mathcal{M}(\mathrm{mt})} \log p_{\theta}(\mathrm{mt}_i \mid \seq_{\setminus i}) - \log p_{\theta}(\mathrm{wt}_i \mid \seq_{\setminus i}),
\end{equation}
where $\mathcal{M}(\mathrm{mt})$ enumerates the mutated positions. Following ProteinGym, we compute the Spearman correlation per assay, average it within each protein and functional category, then across the five categories (Activity, Binding, Expression, Organismal Fitness, Stability). 
No supervision from DMS labels is used.

\paragraph{CASP16 categorical-Jacobian contacts.}

We evaluate residue-residue contact prediction on $41$ CASP16 targets, with a median length of $291$ and a maximum length of $466$, using the categorical-Jacobian construction of \citet{lupo2023categoricaljacobian}. For each position $j$, we substitute each of the $20$ amino acids into the unmasked wild-type sequence and record how the MLM's log-probabilities over the $20$ amino acids at every position $i$ shift relative to the wild type, giving a categorical Jacobian of shape $\seqlen\times20\times\seqlen\times20$. We then center it, symmetrize it, take the Frobenius norm over the amino-acid axes and apply average product correction (APC), and rank residue pairs by the resulting score.
Precision $P$ at the length $T$ of the protein sequence in residues over long-range contacts ($|i-j| \geq 24$) is reported, i.e., $P@T$.

\subsubsection{Supervised Tasks}
\label{app:supervised}

The supervised tasks in our rubric come from the $18$-task suite curated by \citet{chen2024xtrimopglm}, which includes tasks from FLIP~\citep{dallago2021flip} and TAPE~\citep{rao2019evaluating}, among others. These tasks span four biological categories (structure, developability, interaction, and function) and include classification (C) and regression (R) at the residue, sequence, and sequence-pair levels. Examples include secondary structure, fold, fluorescence, stability, GB1 fitness, DeepLoc localization, and solubility prediction. We use the same tasks and metrics with three changes: we evaluate residue-level SSP with both Q$3$ and Q$8$ variants on the pooled test set, omit contact-map prediction because it overlaps with our zero-shot CASP16 contact probe (\S\ref{app:zero_shot}), and replace the original probe with a tuned two-layer MLP head (\S\ref{app:hpo_protocol}). Table~\ref{tab:biomap_tasks} summarizes the resulting $18$ tasks, and Table~\ref{tab:srm_vs_baseline_1m_18tasks} compares baseline AMPLIFY models with our recursive variant after $1$M training steps.

\begin{table}[htb]
    \centering
    \caption{\textbf{The $18$-task supervised suite} from \citet{chen2024xtrimopglm}. ``Type'' distinguishes classification `C' from regression `R' tasks. Sizes are the standard train/test split, where if multiple test splits exist, we report the aggregate size. Validation uses the dataset's own split where one exists and a random $10\%$ of train (seed $42$) otherwise. ``Metric'' follows \citet{chen2024xtrimopglm} and is also the hyperparameter-selection criterion. Results for all $18$ tasks are in Table~\ref{tab:srm_vs_baseline_1m_18tasks}.}
    \label{tab:biomap_tasks}
    \begin{adjustbox}{max width=\linewidth}
        \begin{tabular}{llllll}
            \toprule
            Task & Type & Metric & Prediction target & Source dataset & $|\mathrm{train}|$ / $|\mathrm{test}|$ \\
            \midrule
            SSP Q$3$ & C & acc. & 3-class per-residue secondary structure (H/E/C) & NetSurfP-2.0; test: CB$513$, TS$115$, CASP$12$, CASP$14$ & 10,848 / 667 \\
            SSP Q$8$ & C & acc. & 8-class per-residue secondary structure & NetSurfP-$2.0$; test: CB$513$, TS$115$, CASP$12$, CASP$14$ & 10,848 / 667 \\
            Fold prediction & C & acc. & 1,195-way fold class from SCOP $1.75$ & SCOP / DeepSF split & 12,312 / 3,244 \\
            Solubility prediction & C & acc. & binary soluble vs.\ insoluble & DeepSol & 62,478 / 6,942 \\
            Stability prediction & R & Spearman $\rho$ & protease-concentration fold-retention threshold & TAPE / \citet{rocklin2017massivelyParallel} & 53,614 / 12,851 \\
            Temperature stability & C & MCC & structural stability at $65^\circ\mathrm{C}$ & TemStaPro & 283k / 73k \\
            Optimal temperature & R & Spearman $\rho$ & enzyme optimal reaction temperature & DeepET & 1,706 / 190 \\
            Optimal pH & R & Spearman $\rho$ & enzyme optimal reaction pH & EpHod & 7,124 / 1,971 \\
            Cloning failure & C & AUROC & cloning-stage failure in X-ray crystallography & PredPPCrys & 23,375 / 4,791 \\
            Material production & C & AUROC & material-stage failure in X-ray crystallography & PredPPCrys & 23,339 / 4,791 \\
            Metal-ion binding & C & acc. & presence of any metal-ion binding site & \citet{Cheng2023CoevolutionbasedPO} & 6,000 / 1,332 \\
            Enzyme catalytic efficiency & R & Pearson $r$ & $\log k_\mathrm{cat}$ from the enzyme sequence (no substrate input) & DLKcat & 13,470 / 1,684 \\
            Peptide affinity & C & AUROC & peptide binding to HLA/MHC-I (paired input) & CcBHLA & 57,357 / 8,406 \\
            TCR affinity & C & AUROC & TCR binding to peptide-MHC (paired input) & epiTCR / VDJdb & 19,526 / 4,485 \\
            Fluorescence prediction & R & Spearman $\rho$ & fluorescence intensity of engineered GFP (extrapolation split) & TAPE \citep{rao2019evaluating} & 21,446 / 27,217 \\
            Fitness prediction & R & Spearman $\rho$ & GB1 4-position combinatorial fitness & FLIP GB1 (sampled split) & 6,289 / 1,745 \\
            Localization prediction & C & acc. & 10-way subcellular localization & DeepLoc & 6,622 / 1,842 \\
            Antibiotic resistance & C & acc. & 19-way antibiotic-resistance class & CARD & 2,072 / 1,344 \\
            \bottomrule
        \end{tabular}
    \end{adjustbox}
\end{table}

\paragraph{Why do we report only four of the eighteen supervised tasks?}

\citet{LeBreton2026} find that the aggregate F1/MCC score across the 18 tasks differs by less than $5\%$ among $5$ pLMs (AMPLIFY $120$M/$350$M, ESM-2~$35$M/$150$M/$650$M). To separate genuine pretraining signal from task-intrinsic ceiling effects, they measure the score gap between a shallow head on pretrained embeddings and the same head on randomly initialized weights, averaged over the five models. The results show that only four tasks account for essentially all the pretraining boost: stability prediction ($+57.7\%$), fold prediction ($+44.0\%$), and secondary-structure prediction ($+30.4\%$ Q$3$, $+33.7\%$ Q$8$). Every other task either saturates, with random embeddings already recovering most of the accuracy, or depends on features that MLM captures only weakly. For instance, TCR-pMHC affinity is $25\%$ lower with pretrained embeddings than with random ones. We follow this observation and report the four structural/stability-sensitive tasks (fold, SSP mean of Q$3$ and Q$8$, stability, and localization) in the main-paper Table~\ref{tab:srm_vs_baseline_1m} and Appendix Table~\ref{tab:srm_vs_baseline_1m_std}. We further detail these tasks below.

\paragraph{Fold prediction.}

Given an amino-acid sequence, the model must predict its top-1 fold class out of $1{,}195$ SCOP $1.75$ folds. We use the DeepSF split with $12{,}312$ train and $3{,}244$ test samples. Because remote homologs are grouped into the same fold, high accuracy requires representations that recognize structural similarity across sequences with $<25\%$ identity. This is exactly the regime where handcrafted alignment features fail and where pretraining should help. The metric is the sequence-level total accuracy (top-1 fold match)~\citep{rao2019evaluating, Fox2013SCOPeSC}.

\paragraph{Localization prediction.}

A $10$-way sequence-level classification of subcellular localization derived from UniProt annotations of $\sim 8{.}5$k eukaryotic proteins with $6{,}622$ train and $1{,}842$ test samples~\citep{Armenteros2017deeploc}. Signal peptides and transmembrane spans are informative, but the label ultimately requires integrating long-range motifs and organelle-specific sequence patterns, which frozen bag-of-embedding baselines only partially capture. We report the total test accuracy here.

\paragraph{Stability prediction.}

A sequence-level regression on the protease concentration at which a designed mini-protein retains its native fold, from the massively parallel proteolysis assay of \citet{rocklin2017massivelyParallel} with $53{,}614$ train and $12{,}851$ test sequences. The target is a real-valued proxy for thermodynamic stability, and the reported score is the Spearman rank correlation between predicted and measured values across the test split. \citet{LeBreton2026} find stability to be the single largest pretraining-driven task where pretraining performs up to $+57.7\%$ over the random baseline. This makes it the most discriminative task for comparing pLMs at similar overall scale.

\paragraph{Secondary-structure prediction (SSP).}

A per-residue classification into either $3$ classes (Q$3$ with helix, strand and coil) or $8$ classes (Q$8$ with DSSP labels) trained on the NetSurfP-2.0 corpus~\citep{Klausen2018NetSurfP20IP} and evaluated on the pooled test split ($667$ proteins from CB$513$, TS$115$, CASP$12$ and CASP$14$), with per-residue accuracy computed over all scored test residues. Q$3$ and Q$8$ use the same encoder feature map and differ only in the final softmax head. We report the equal-weight mean of the two per-residue accuracies. Secondary structure is the residue-level task where pretraining most clearly beats random embeddings~\citep{LeBreton2026} because the label is a direct function of local backbone geometry that MLM implicitly learns during masked reconstruction.

\subsection{Task Head and Hyperparameter Search for Downstream Supervised Tasks}
\label{app:hpo_protocol}

During evaluation of the downstream supervised tasks, the encoder is frozen. Embeddings are extracted once and cached: the recurrent-core state for fixed- and variable-depth recAMPLIFY and the coda output for GMR variants and the final-layer output for AMPLIFY and ESM-2. Sequence-level tasks mean-pool over residue positions, excluding the special tokens. On top of these embeddings, we fit a two-layer MLP head (one hidden GELU layer) with a softmax output for classification and a single-scalar linear output for regression. Regression targets are standardized with the training set's mean and standard deviation. The head is selected by a $50$-trial Bayesian TPE search over hidden size $\{1024, 2048\}$, learning rate $[10^{-4}, 10^{-3}]$, number of epochs $\{10, 20, 40\}$, weight decay $[10^{-4}, 10^{-3}]$ and batch size $\{32, 64, 128\}$. For SSP, we instead search the corresponding exhaustive grid of $72$ configurations. Each configuration is trained with early stopping on the task's validation split (the dataset's own split where provided, otherwise a random $10\%$ of the training set), and selection uses the task's reported metric. The best configuration is then retrained with $5$ random seeds and evaluated on the held-out test split. We report the mean (main-paper Table~\ref{tab:srm_vs_baseline_1m}) and the standard deviation (Appendix Table~\ref{tab:srm_vs_baseline_1m_std}).

\section{Additional Results}
\begin{table}[htb]
    \centering
    \caption{\textbf{Mean $\pm$ standard deviation scores for supervised probes} reported by rerunning the best HPO configuration over $5$ seeds. Small standard deviations signify that most of the gains from recursion are beyond the noise regime.}
    \label{tab:srm_vs_baseline_1m_std}
    \begin{adjustbox}{max width=\linewidth}
        \begin{tabular}{lcccc}
            \toprule
            Model & Fold\,$\uparrow$ & Localization\,$\uparrow$ & Stability\,$\rho$\,$\uparrow$ & SSP\,$\uparrow$ \\
            \midrule
            \multicolumn{5}{c}{$\mathbf{1}$\textbf{M pretraining steps}} \\
            \midrule
            AMPLIFY-$120$M ($1$M) & 0.566~$\pm$~0.006 & 0.651~$\pm$~0.008 & 0.591~$\pm$~0.008 & 0.724~$\pm$~0.001 \\
            AMPLIFY-$350$M ($1$M) & 0.568~$\pm$~0.004 & 0.716~$\pm$~0.005 & 0.585~$\pm$~0.020 & 0.742~$\pm$~0.000 \\
            recAMPLIFY ($1$M) & 0.696~$\pm$~0.002 & 0.741~$\pm$~0.001 & 0.703~$\pm$~0.025 & 0.742~$\pm$~0.000 \\
            \midrule
            \multicolumn{5}{c}{$\mathbf{100}$\textbf{K pretraining steps}} \\
            \midrule
            AMPLIFY-$120$M ($100$K) & 0.602~$\pm$~0.002 & 0.691~$\pm$~0.005 & 0.611~$\pm$~0.048 & 0.720~$\pm$~0.000 \\
            AMPLIFY-$350$M ($100$K) & 0.570~$\pm$~0.005 & 0.672~$\pm$~0.005 & 0.479~$\pm$~0.027 & 0.729~$\pm$~0.000 \\
            recAMPLIFY (fixed-L) & 0.601~$\pm$~0.000 & 0.700~$\pm$~0.006 & 0.644~$\pm$~0.002 & 0.720~$\pm$~0.001 \\
            recAMPLIFY (varying-L) & 0.653~$\pm$~0.003 & 0.711~$\pm$~0.005 & 0.688~$\pm$~0.060 & 0.726~$\pm$~0.001 \\
            $+$GMR (halt-only) & 0.637~$\pm$~0.002 & 0.714~$\pm$~0.008 & 0.650~$\pm$~0.027 & 0.729~$\pm$~0.001 \\
            $+$GMR (scalar gate) & 0.628~$\pm$~0.001 & 0.708~$\pm$~0.006 & 0.737~$\pm$~0.015 & 0.722~$\pm$~0.000 \\
            $+$GMR (rank-$16$ gate) & 0.621~$\pm$~0.006 & 0.708~$\pm$~0.006 & 0.703~$\pm$~0.013 & 0.721~$\pm$~0.000 \\
            $+$GMR (per-dim gate) & 0.621~$\pm$~0.004 & 0.718~$\pm$~0.005 & 0.627~$\pm$~0.037 & 0.722~$\pm$~0.000 \\
            \bottomrule
        \end{tabular}
    \end{adjustbox}
\end{table}

\begin{table}[htb]
    \centering
    \caption{\textbf{Throughput comparison for our recursive pretraining recipe} at three cycle budgets $\ncyc$ against that of the non-recurrent AMPLIFY baselines. ``Blocks'' is the effective depth $\effdepth = p + \ncyc r + c$ used in Table~\ref{tab:srm_vs_baseline_1m}, i.e., prelude, $\ncyc$ applications of the tied recurrent core, and coda. The $133$M recAMPLIFY rows use $(p,r,c) = (2, 8, 2)$. The last column normalizes the throughput by that count. All reported values are measured over $500$ optimizer steps on one A$100$-SXM$4$-$40$GB with $32$ sequences per device packed into a single \textsc{FlashAttention} varlen buffer ($\approx10.9$K tokens) using \texttt{bf16} precision without gradient accumulation. Throughput is logged over every 100-step window. To compensate for the warm-up from data loading in the early windows, we use the mean of the last two windows for comparisons.}
    \label{tab:cost}
    \begin{adjustbox}{max width=\linewidth}
        \begin{tabular}{lrrrrrr}
            \toprule
            & \multicolumn{3}{c}{Model size} & \multicolumn{3}{c}{Measured training cost} \\
            \cmidrule(lr){2-4}\cmidrule(lr){5-7}
            Model
            & \makecell[r]{Trainable\\parameters}
            & \makecell[r]{Hidden\\width $\dmodel$}
            & \makecell[r]{Blocks per\\forward pass}
            & \makecell[r]{Throughput\\(K tokens/s)}
            & \makecell[r]{Relative\\throughput}
            & \makecell[r]{Throughput per block\\(M blk$\cdot$tok/s)} \\
            \midrule
            AMPLIFY-$120$M & $118$M & $640$ & $24$ & $142.4$ & $1.00\times$ & $3.42$ \\
            AMPLIFY-$350$M & $354$M & $960$ & $32$ & $59.1$ & $0.41\times$ & $1.89$ \\
            \midrule
            recAMPLIFY (fixed $\ncyc=3$) & $133$M & $960$ & $28$ & $54.3$ & $0.38\times$ & $1.52$ \\
            recAMPLIFY (fixed $\ncyc=5$) & $133$M & $960$ & $44$ & $37.9$ & $0.27\times$ & $1.67$ \\
            recAMPLIFY (fixed $\ncyc=7$) & $133$M & $960$ & $60$ & $29.1$ & $0.20\times$ & $1.75$ \\
            \bottomrule
        \end{tabular}
    \end{adjustbox}
\end{table}

\begin{table}[htb]
    \centering
    \caption{\textbf{Full $18$-task supervised comparison at $1$M pretraining steps.} Each cell reports the mean $\pm$ standard deviation of the test-set metric for the best HPO configuration retrained with $5$ random seeds (\S\ref{app:hpo_protocol}). ``Metric'' follows the source benchmark (Table~\ref{tab:biomap_tasks}): accuracy, MCC or AUROC for classification and Spearman or Pearson correlation for regression; higher is better for every metric. AMPLIFY-$120$M and AMPLIFY-$350$M are the standard AMPLIFY $1$M-step checkpoints and recAMPLIFY is our recursive $1$M checkpoint with $(p,r,c)=(2,10,2)$ that is trained and evaluated at $\ncyc_{\mathrm{eval}} = \ncyc = 4$. Best cell per task is \textbf{bold}. recAMPLIFY wins on $14$ of $18$ tasks despite having $56\%$ fewer unique parameters than AMPLIFY-$350$M.}
    \label{tab:srm_vs_baseline_1m_18tasks}
    \begin{adjustbox}{max width=\linewidth}
        \begin{tabular}{llccc}
            \toprule
            Task & Metric & AMPLIFY-$120$M & AMPLIFY-$350$M & recAMPLIFY \\
            \midrule
            SSP Q$3$ & acc. & 0.793~$\pm$~0.001 & 0.808~$\pm$~0.001 & \textbf{0.810~$\pm$~0.001} \\
            SSP Q$8$ & acc. & 0.655~$\pm$~0.000 & \textbf{0.677~$\pm$~0.000} & 0.673~$\pm$~0.001 \\
            Fold prediction & acc. & 0.566~$\pm$~0.006 & 0.568~$\pm$~0.004 & \textbf{0.696~$\pm$~0.002} \\
            Solubility prediction & acc. & 0.690~$\pm$~0.011 & \textbf{0.718~$\pm$~0.004} & 0.711~$\pm$~0.002 \\
            Stability prediction & Sp. $\rho$ & 0.591~$\pm$~0.008 & 0.585~$\pm$~0.020 & \textbf{0.703~$\pm$~0.025} \\
            Temperature stability & MCC & 0.837~$\pm$~0.005 & \textbf{0.868~$\pm$~0.006} & 0.852~$\pm$~0.004 \\
            Optimal temperature & Sp. $\rho$ & 0.577~$\pm$~0.002 & \textbf{0.653~$\pm$~0.005} & 0.534~$\pm$~0.020 \\
            Optimal pH & Sp. $\rho$ & 0.491~$\pm$~0.003 & 0.520~$\pm$~0.003 & \textbf{0.594~$\pm$~0.013} \\
            Cloning failure & AUROC & 0.768~$\pm$~0.005 & 0.774~$\pm$~0.007 & \textbf{0.807~$\pm$~0.002} \\
            Material production & AUROC & 0.843~$\pm$~0.001 & 0.847~$\pm$~0.001 & \textbf{0.855~$\pm$~0.002} \\
            Metal-ion binding & acc. & 0.710~$\pm$~0.012 & 0.733~$\pm$~0.004 & \textbf{0.793~$\pm$~0.018} \\
            Enzyme catalytic efficiency & Pearson $r$ & 0.585~$\pm$~0.003 & 0.640~$\pm$~0.000 & \textbf{0.648~$\pm$~0.034} \\
            Peptide affinity & AUROC & 0.723~$\pm$~0.001 & 0.820~$\pm$~0.002 & \textbf{0.928~$\pm$~0.001} \\
            TCR affinity & AUROC & 0.643~$\pm$~0.015 & 0.754~$\pm$~0.000 & \textbf{0.904~$\pm$~0.001} \\
            Fluorescence prediction & Sp. $\rho$ & 0.239~$\pm$~0.031 & 0.236~$\pm$~0.003 & \textbf{0.508~$\pm$~0.006} \\
            Fitness prediction & Sp. $\rho$ & 0.415~$\pm$~0.050 & 0.462~$\pm$~0.029 & \textbf{0.713~$\pm$~0.000} \\
            Localization prediction & acc. & 0.651~$\pm$~0.008 & 0.716~$\pm$~0.005 & \textbf{0.741~$\pm$~0.001} \\
            Antibiotic resistance & acc. & 0.940~$\pm$~0.015 & 0.947~$\pm$~0.008 & \textbf{0.967~$\pm$~0.003} \\
            \bottomrule
        \end{tabular}
    \end{adjustbox}
\end{table}
\begin{figure}[htb]
    \centering
    \includegraphics[width=0.95\textwidth]{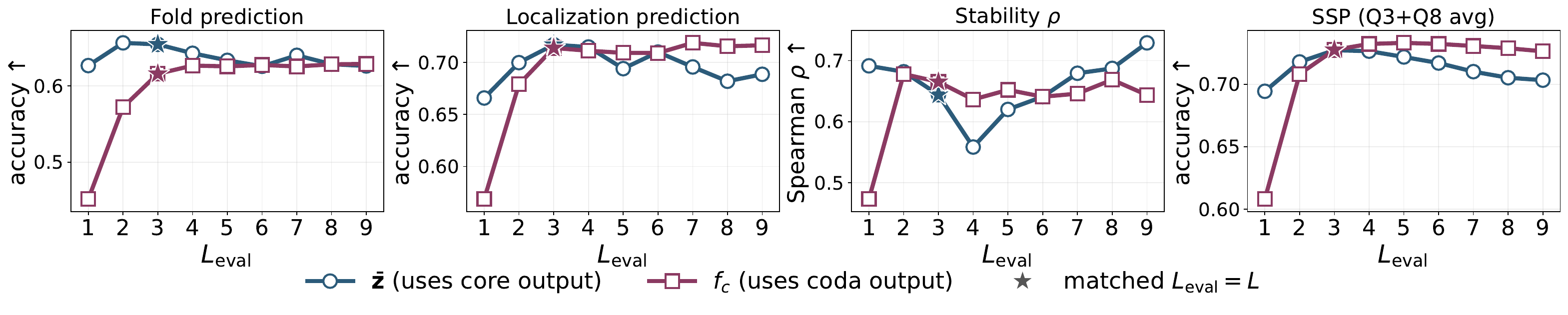}
    \caption{\textbf{Performance comparison of different representation choices used for training downstream supervised task probes:} $z$ uses the output of the recurrent core while \emph{coda} is the standard pre-LM head output $f_c(\widetilde{\mathbf z})$. We sweep $\ncyc_{\mathrm{eval}}\in\{1,\dots,9\}$ on the same recAMPLIFY trained with $(p,r,c) = (2,8,2)$, and $\ncyc = 3$ ($\star$ denotes $\ncyc_{\mathrm{eval}}$ with matched $\ncyc$).}
    \label{fig:vary_embed_mode}
\end{figure}

\paragraph{Recursion offers a better representation choice for training supervised probes.}

Besides the standard coda output $f_c(\widetilde{\mathbf z})$, recAMPLIFY exposes a second readout choice for training downstream supervised task probes: the refined latent state $\bar{\mathbf{z}}$. Figure~\ref{fig:vary_embed_mode} shows that the latent state $\bar{\mathbf{z}}$ provides a better representation space for most supervised tasks at the depths seen during training ($\ncyc_{\mathrm{eval}} \le 3$), with the largest margins at shallow depth (e.g., at $\ncyc_{\mathrm{eval}} = 1$, fold $0.63$ vs.\ $0.45$, stability $0.69$ vs.\ $0.47$). In the unseen depth regime ($\ncyc_{\mathrm{eval}} > 3$), coda probes are steadier on SSP and localization, where $\bar{\mathbf{z}}$ gradually declines, while the two are on par for fold and $\bar{\mathbf{z}}$ recovers and overtakes coda on stability at the largest depths. Given its superior performance for seen recurrence depths, we report supervised probe results for fixed- and variable-depth recAMPLIFY using the latent state $\bar{\mathbf z}$. Our GMR variants are probed on their coda output, which decodes the halting mixture.

\begin{figure}[htb]
    \centering
    \begin{subfigure}[t]{0.9\textwidth}
        \centering
        \includegraphics[width=\linewidth]{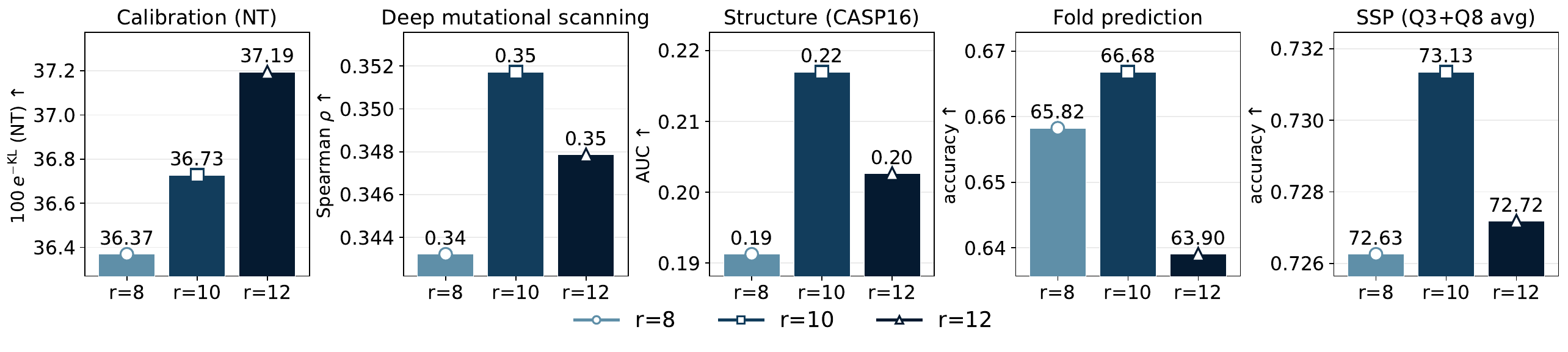}
        \caption{Varying recursive layers $r$ while fixing $(p,c)=(2,2)$.}
        \label{fig:vary_r}
    \end{subfigure}

    \begin{subfigure}[t]{0.9\textwidth}
        \centering
        \includegraphics[width=\linewidth]{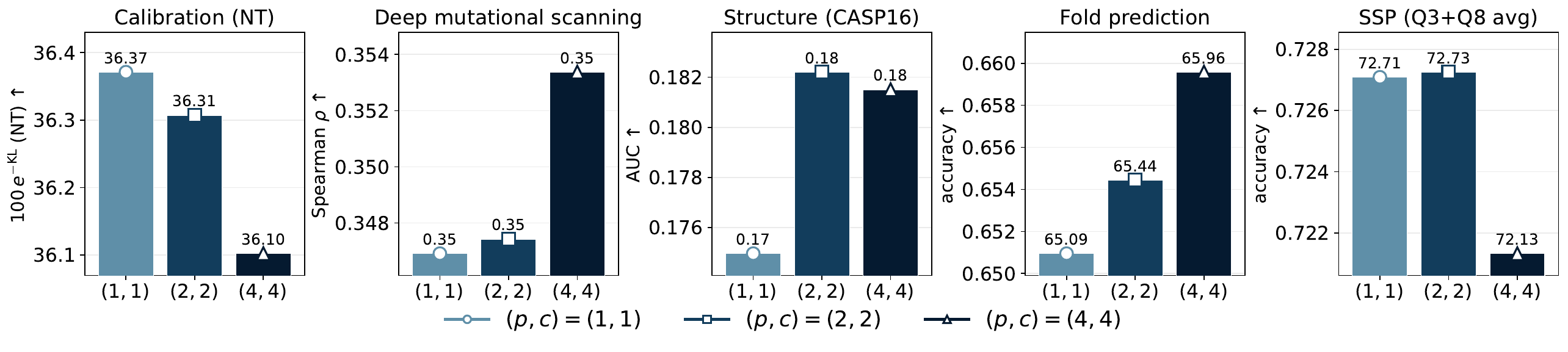}
        \caption{Varying the prelude $p$ and coda $c$ while adjusting $r$ to keep the number of layers $p + r + c = 12$.}
        \label{fig:vary_prelude_coda}
    \end{subfigure}
    \caption{\textbf{Ablation of physical layer counts:} we train each variant for $100$K steps and evaluate at its training depth: $L=L_{\mathrm{eval}}=4$ in (a) and $L=L_{\mathrm{eval}}=3$ in (b). Overall, $r=10$ performs best in (a) on DMS, CASP, fold and SSP, while $r=12$ is best on KL (NT). The prelude/coda split in (b) changes every probe by less than $1$ pp.}
\end{figure}

\paragraph{Design decisions that do not scale well for pretraining.}

\begin{figure}[htb]
    \centering
    \includegraphics[width=\textwidth]{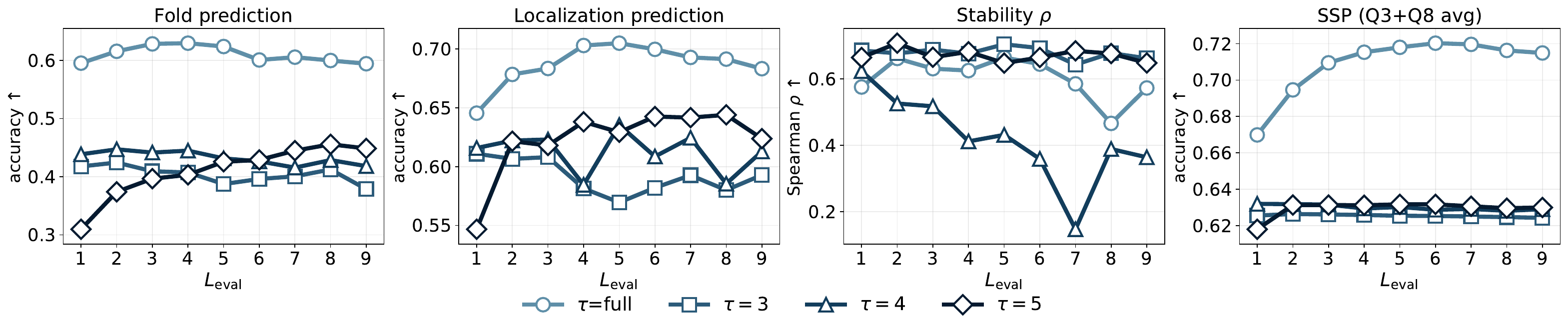}
    \caption{\textbf{Effect of truncated backpropagation (BPTT) steps on supervised probes.} We sweep across $L_{\mathrm{eval}}\in\{1,\dots,9\}$ for four checkpoints that are trained with $L=6$ but differ in the truncated backprop-through-time window $\tau$: full BPTT ($\tau=$full, gradients through every cycle) and truncated windows $\tau\in\{3,4,5\}$. Here, the last $\tau$ cycles receive gradients, while the earlier $L-\tau$ do not. Full-BPTT holds up across the sweep and remains competitive with AMPLIFY-$350$M (main-paper Table~\ref{tab:srm_vs_baseline_1m}). Truncated variants lose about $0.2$ in fold accuracy and $0.08$ in SSP accuracy at every depth, and do not benefit from additional cycles on SSP. Stability is the exception, with $\tau = 3$ and $\tau = 5$ matching or exceeding full BPTT, while $\tau = 4$ becomes unstable across seeds at larger depths.}
    \label{fig:vary_tbtt_new}
\end{figure}

\begin{figure}[htb]
    \centering
    \includegraphics[width=\textwidth]{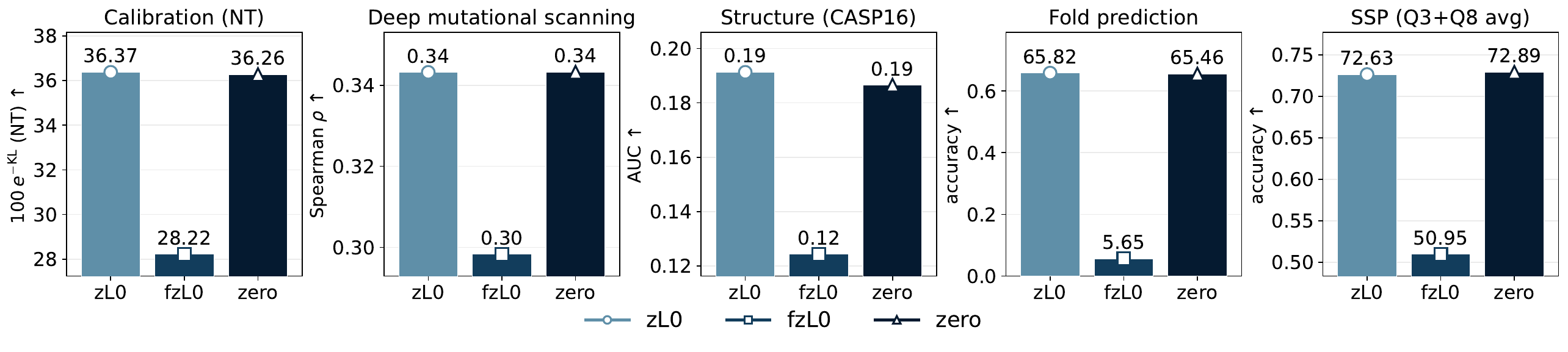}
    \caption{\textbf{Varying context type for $L = 4$.} All models are evaluated at the matched $L_{\mathrm{eval}}=4$ depth~\citep{srm}. The simplest context type, `zero', which uses an identity transformation of the previous cycle's latent $z_{k-1}$, performs surprisingly well throughout all tasks.}
    \label{fig:vary_context_type}
\end{figure}

Inspired by a recent line of work on recursive refinement in a limited-data regime, we ablate the effects of truncated backpropagation (BPTT)~\citep{trm} and context type~\citep{srm}. Figures~\ref{fig:vary_tbtt_new} and~\ref{fig:vary_context_type} show that neither design helps at pretraining scale: truncated BPTT degrades fold, localization and SSP at every evaluation depth, and a learned context transform (fzL0) collapses, whereas the identity (`zero') and zL0 contexts perform comparably. We therefore train with full BPTT and an identity transformation of the previous cycle's latent state.

\section{Additional Mechanistic Analyses}
\label{app:sec_mech_analyses}

\paragraph{Gates help achieve stable attention geometry.}

For the recurrent block to converge to an actual fixed point, the attention pattern it induces must also stabilize in addition to the hidden states. Figure~\ref{app:fig_mech_post_halt_attn} shows that once past the trained horizon ($\ncyc>6$), the scalar gate's cross-cycle attention-map distances collapse, whereas the variable-depth model at fixed test-time depth and halt-only GMR retain cycle-dependent attention maps alongside their continued state drift. The low-rank gate's attention distances also fail to collapse, which is consistent with its magnitude drift.

\begin{figure}[htb]
    \centering
    \includegraphics[width=0.85\textwidth]{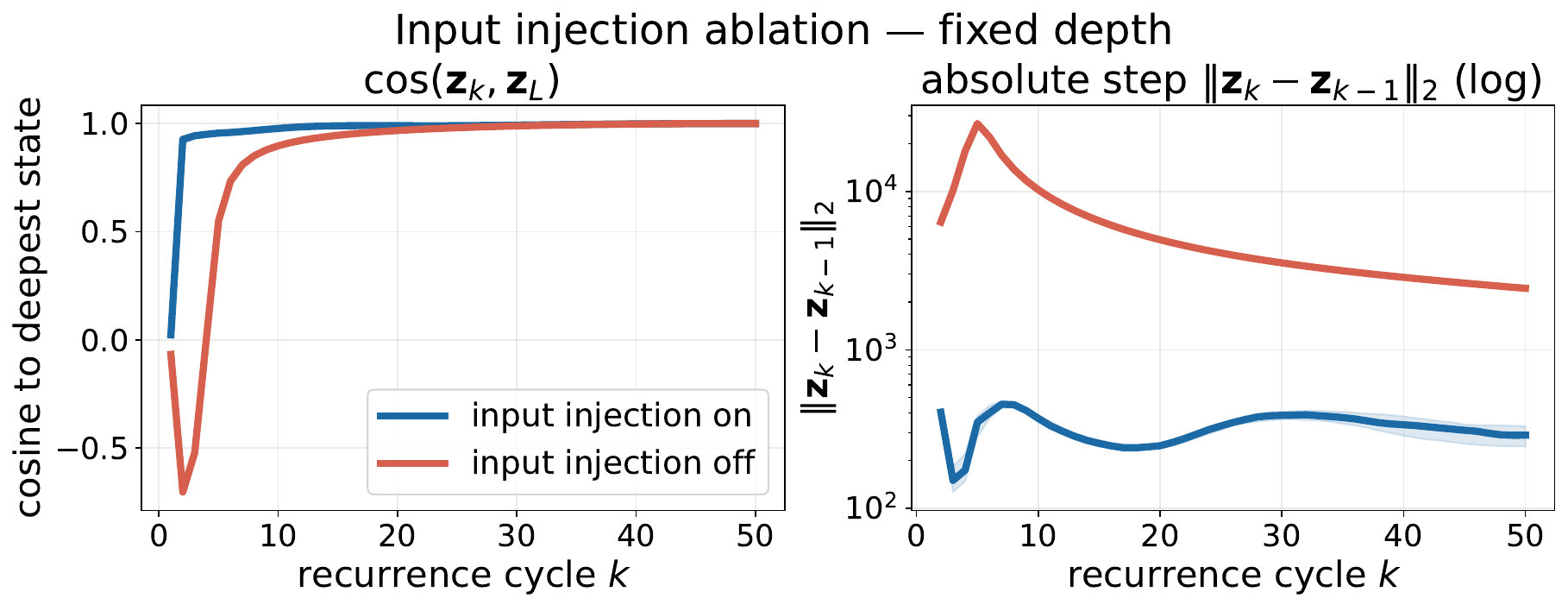}
\caption{\textbf{Input-injection ablation} for recAMPLIFY. Blue line shows a normal forward pass while red denotes injection stream removed from the recurrent core. \textbf{Left:} $\cos(\mathbf{z}_\kcyc,\mathbf{z}_L)$. \textbf{Right:} $\|\mathbf{z}_\kcyc{-}\mathbf{z}_{\kcyc-1}\|_2$ on log-$y$. With input injection, the state is a bounded iterated map ($\|\Delta\mathbf{z}\|_2$ plateaus below $10^3$). Without the injection, the state diverges by three orders of magnitude within a few cycles. Input injection thus helps keep the recurrence bounded, matching the finding of  \citet{blayney2026mechanistic} on causal-LM looped Transformers. We note that this boundedness does not imply convergence. The blue plateau in the right panel stays around $\|\Delta\mathbf{z}\|_2 \approx 300$ throughout the OOD range, motivating main-paper Figure~\ref{fig:mech_state_convergence}.}
    \label{app:fig_mech_injection}
\end{figure}

\begin{figure}[htb]
    \centering
    \begin{subfigure}{0.95\linewidth}
        \centering
        \includegraphics[width=\linewidth]{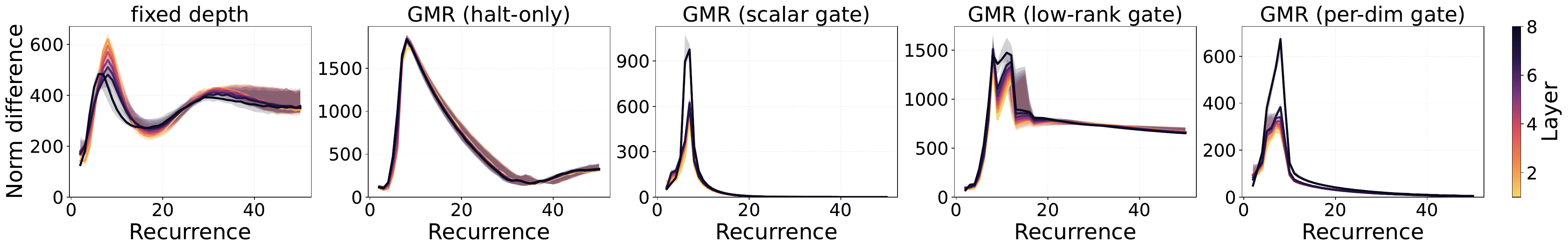}
        \caption{Difference across successive iterations $\|\mathbf{z}^{\ell}_{\kcyc}-\mathbf{z}^{\ell}_{\kcyc-1}\|_2$.}
    \end{subfigure}

    \begin{subfigure}{0.95\linewidth}
        \centering
        \includegraphics[width=\linewidth]{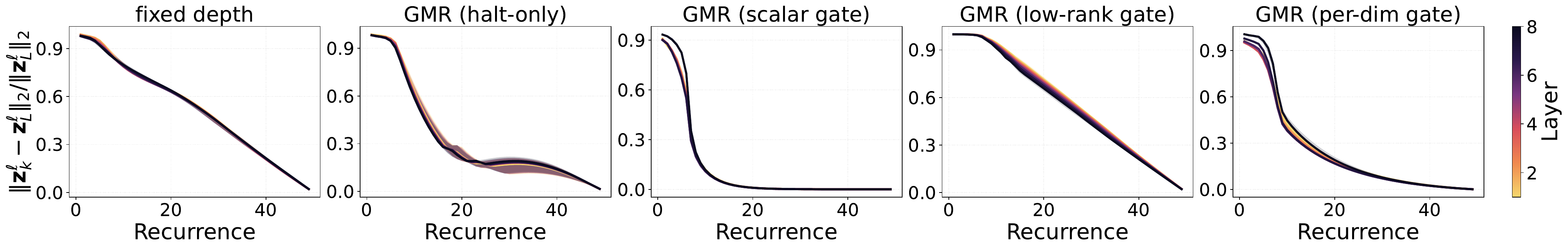}
        \caption{Normalized distance to the deepest state.}
    \end{subfigure}

    \begin{subfigure}{0.95\linewidth}
        \centering
        \includegraphics[width=\linewidth]{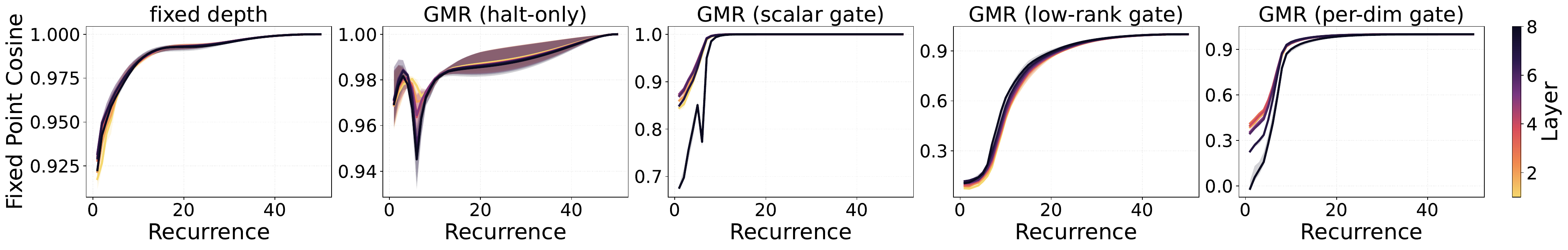}
        \caption{Cosine similarity $\cos(\mathbf{z}^{\ell}_{\kcyc},\mathbf{z}^{\ell}_L)$ with the deepest recorded state.}
    \end{subfigure}
    \caption{\textbf{Per-layer dynamics of each recurrent block layer $\ell$ on the stability prediction task} (median over $32$ probes). (a) Successive-difference norm: gating drives each layer's step size toward zero past the $\ncyc = 6$ cycles seen during training. (b) Normalized operator residual to each layer's deepest recorded state $\mathbf{z}^{\ell}_L$ (taken as an approximate fixed point) over $\kcyc\in\{1,\ldots,L - 1\}$ cycles shows the key diagnostic being the \emph{rate} and \emph{timing} of decay. Scalar and per-dim gates reach near-zero residual within a few cycles of the trained horizon, whereas halt-only and rank-$16$ decay only slowly and remain substantially non-zero at intermediate cycles. (c) Cosine similarity to $\mathbf{z}^{\ell}_L$: alignment is near-perfect for every variant, so cosine is an \emph{indirect} diagnostic and must be read jointly with (a) and (b). Read together, only the scalar and per-dimension gates deliver empirical fixed-point convergence.}
    \label{fig:mech_state_convergence_full}
\end{figure}

\begin{figure}[htb]
    \centering
    \includegraphics[width=\textwidth]{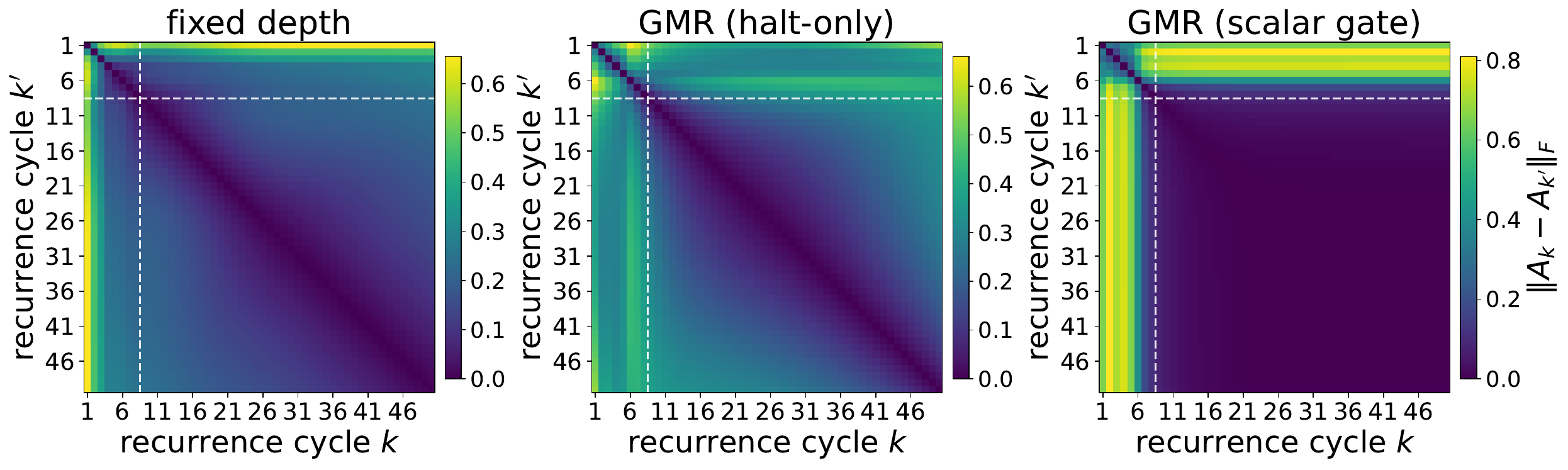}
    \caption{\textbf{Post-halt attention stability.} Head-averaged attention Frobenius distance between recurrent cycles for three representative variants at $\ncyc_{\mathrm{eval}} = 50$. Each cell $(\kcyc,\kcyc')$ is $\|A_\kcyc-A_{\kcyc'}\|_F$ averaged over $B = 4$ probe sequences from the stability prediction task, and the dashed line marks the trained halt horizon at $\kcyc \approx 7$. For the variant with fixed test-time depth and varying-$\ncyc$ pretraining (left), attention keeps drifting through all 50 cycles, despite the recurrence being bounded (Figure~\ref{app:fig_mech_injection}). For halt-only GMR (middle), a discontinuity appears at the halt horizon but off-diagonal cells continue to darken smoothly. For the scalar gate, cycles $8$--$50$ collapse into uniform differences. The comparison isolates the role of the residual gate where neither variable-$\ncyc$ training nor a halt signal on its own produces a post-halt attention fixed point; only the gate achieves this.}
    \label{app:fig_mech_post_halt_attn}
\end{figure}

\paragraph{ColSum concentration.}
\label{app:colsum_definition}

For a row-stochastic attention matrix $A$ of a sequence with real-token length $\seqlen$, we define the column mass $c_j = \sum_i A_{ij}$ and its length-normalized version $\hat c_j = c_j / \seqlen$. The ColSum concentration is then given as:
\begin{equation}
    C = 1 + \frac{1}{\log \seqlen}\, \sum_j \hat c_j \log \hat c_j \in [0, 1],
\end{equation}
where the right-hand-side term is the residual of the length-normalized entropy of the column-mass distribution. $C = 0$ therefore corresponds to uniform mixing, where every key receives an equal share of the attention mass and $C = 1$ corresponds to all mass pooled on a single key. We compute $C$ per attention head, per protein, and per physical layer, and report the mean over heads and over the $32$ probe proteins of our stability probe. Query positions are already summed inside $\hat c_j$ so no averaging over query indices is needed. The dashed red curve throughout \S\ref{sec:stages_of_inference} is our \emph{feedforward reference} consisting of a non-recurrent recAMPLIFY with identical width, attention heads, and $p + r + c=12$ total layers, trained at $\ncyc=1$ so the core layers are applied untied. A non-flat layerwise profile of $C$ signals mixing heterogeneity across physical layers while proximity to the dashed red curve signals fidelity to the feedforward reference in both shape and amplitude.

\begin{figure}[htb]
    \centering
    \includegraphics[width=0.95\linewidth]{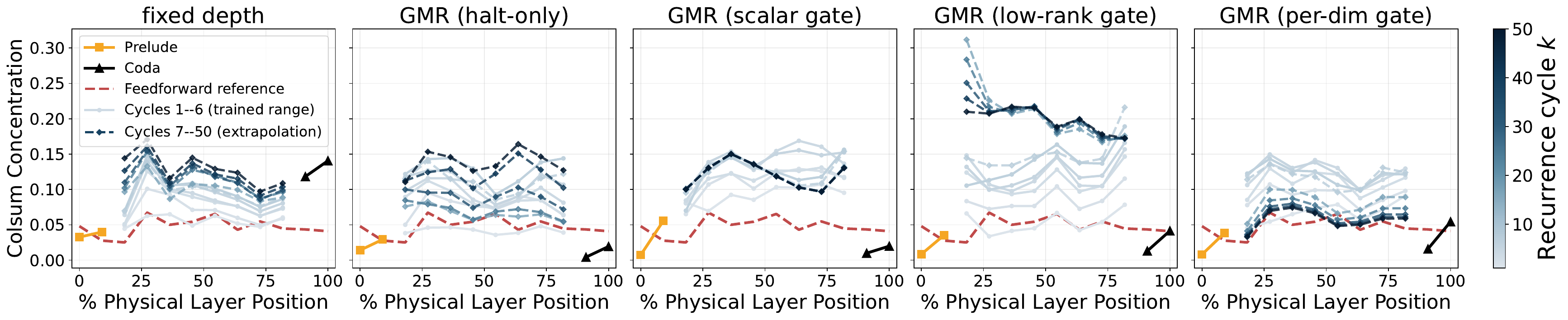}
    \caption{ColSum concentration vs.\ physical layer position across five recAMPLIFY variants, evaluated at $L_\mathrm{eval}=50$. Solid lines with circles are the trained cycles ($[1,6]$). Dashed lines with diamonds: the six evenly spaced samples from the extrapolation range $[7,50]$, orange square: prelude, black triangle: coda. The dashed red line is the feedforward reference which is a recAMPLIFY trained with the same $(p,r,c)=(2,8,2)$ physical layer counts without recursion, i.e., $L=1$.}
    \label{fig:mech_stages_of_inference}
\end{figure}

\section{Additional Results on GMR Depth Encoding Biological Signal}
\label{sec:biological_properties}

\begin{table}[t]
    \centering
    \caption{Partial correlations between GMR halt depth and sharpness across probing tasks alongside the compute budget: \textbf{raw} $=$ $|r(\En,\mathrm{err})|\,|\,\log \seqlen$ denotes the partial correlation between per-protein expected halt depth $\En$ and task-specific per-protein error $\mathrm{err}$, after regressing out the log sequence length $\log \seqlen$; \textbf{ctrl} $=$ $|r(\En,\mathrm{err})|\,|\,\log \seqlen{,}\log\mathrm{Sharp}$ denotes the same after additionally regressing out the task-specific sharpness proxy $\log\mathrm{Sharp}$; \textbf{coup} $=$ $r(\En,\log\mathrm{Sharp})\,|\,\log \seqlen$ is the signed length-controlled partial correlation of $\En$ with $\log\mathrm{Sharp}$. Per-task error and sharpness proxy: for KL, $\mathrm{err}$ is the mean $\KL(\mathbf f_i|p_\theta)$ and sharpness is the model's mean prediction sharpness $-H[p_\theta]$; for fold and localization, $\mathrm{err}$ is test misclassification and sharpness is the negative softmax entropy of the classifier; for DMS, $\mathrm{err}=1-\rho$ per assay and sharpness is the log standard deviation of the model's scores. The two rightmost columns give per-protein $\En$ (on the KL probe) and the potential budget reduction $1 - \En/\ncyc$ with respect to the fixed $\ncyc = 6$ budget under hard stopping.}
    \label{tab:halt_sharpness_multitask}
    \begin{adjustbox}{max width=\linewidth}
        \begin{tabular}{lccc|ccc|ccc|ccc|cc}
            \toprule
            & \multicolumn{3}{c|}{KL calibration}
            & \multicolumn{3}{c|}{Fold prediction}
            & \multicolumn{3}{c|}{Localization prediction}
            & \multicolumn{3}{c|}{ProteinGym DMS}
            & \multicolumn{2}{c}{Compute} \\
            Variant & raw & ctrl & coup & raw & ctrl & coup & raw & ctrl & coup & raw & ctrl & coup & $\En$ & reduction \\
            \midrule
            Halt-only GMR & $0.42$ & $0.11$ & $+0.50$ & $0.08$ & $0.01$ & $-0.15$ & $0.08$ & $0.04$ & $-0.13$ & $0.13$ & $0.08$ & $+0.10$ & $3.57$ & $40\%$ \\
            Scalar gate & $0.28$ & $0.03$ & $+0.29$ & $0.11$ & $0.04$ & $-0.13$ & $0.07$ & $0.02$ & $-0.16$ & $0.06$ & $0.01$ & $+0.10$ & $4.07$ & $32\%$ \\
            Low-rank-$16$ & $\best{0.59}$ & $0.05$ & $\best{+0.63}$ & $0.06$ & $0.02$ & $+0.08$ & $0.02$ & $0.00$ & $-0.08$ & $0.00$ & $0.04$ & $+0.05$ & $4.06$ & $32\%$ \\
            Per-dim & $0.21$ & $0.04$ & $+0.22$ & $0.00$ & $0.02$ & $-0.03$ & $0.03$ & $0.01$ & $-0.06$ & $0.05$ & $0.04$ & $-0.04$ & $4.08$ & $32\%$ \\
            \bottomrule
        \end{tabular}
    \end{adjustbox}
\end{table}

\paragraph{Compositional axes of the GMR halt signal.}
\label{app:halt_composition}

\begin{figure}[h]
    \centering
    \includegraphics[width=0.8\linewidth]{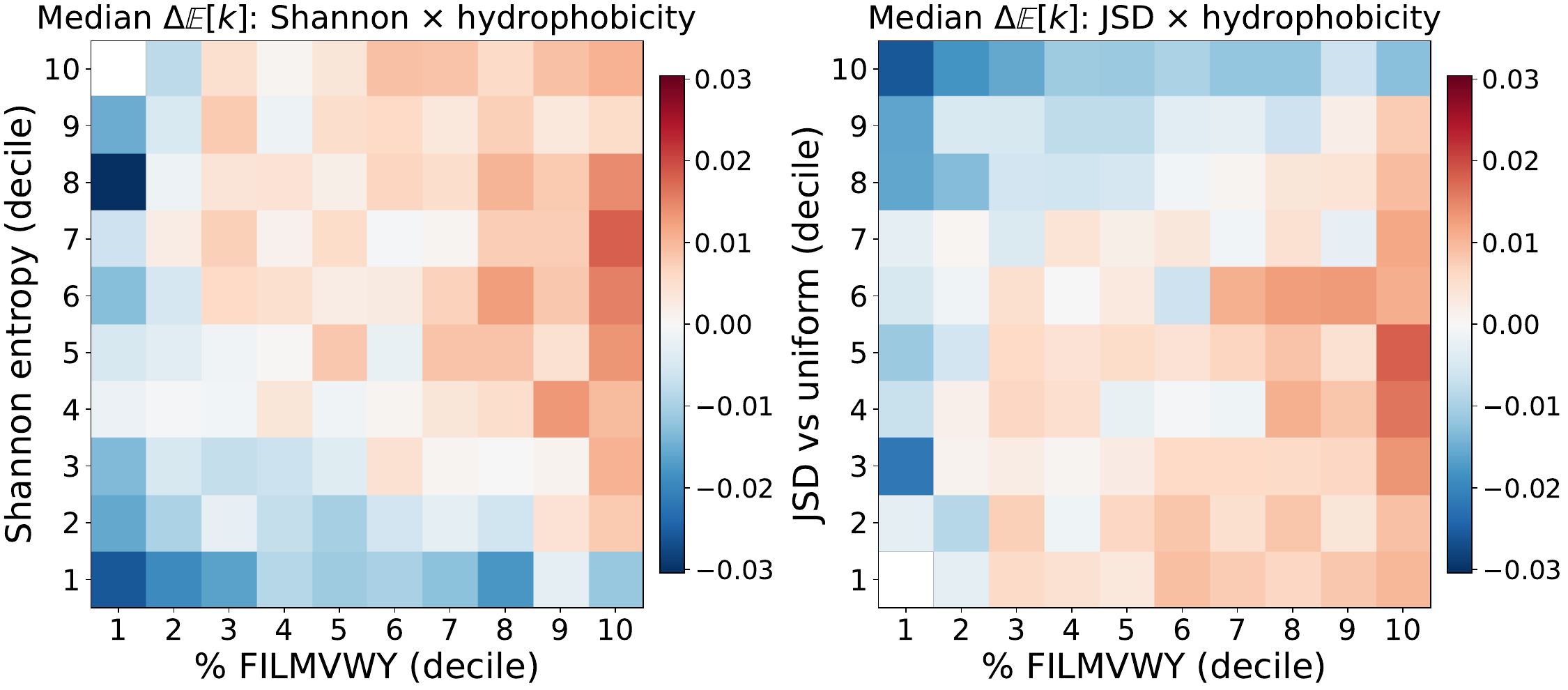}
    \caption{\textbf{Median residual halt depth $\En$ binned by composition-diversity $\times$ hydrophobicity.} Left: Shannon entropy of the AA distribution (rows) against $\%$ of $\text{FILMVWY}$ (columns). Right: Jensen--Shannon divergence of the AA distribution to the uniform $1/20$ prior (rows) against $\%$ of $\text{FILMVWY}$ (columns). Both panels are computed on the $10$K UniProt held-out sequences of \S\ref{sec:halt_biology} and use $\En$ residualized against $\log T$ and the model's prediction sharpness $-H[p_\theta]$. Diverging color scale is centered at $0$. Positive (red) means the model spends more cycles than a length- and sharpness-matched null.}
    \label{app:fig_halt_comp_diversity}
\end{figure}

Main-paper Table~\ref{tab:cross_task_property_signatures} shows two independent per-protein composition signals for the controlled GMR halt depth $\En$ (residualized on $\log T$ and $-H[p_\theta]$): high sequence entropy and high hydrophobic (FILMVWY) content both correlate with depth. We now show that these are two projections of a single compositional-vocabulary axis. Figure~\ref{app:fig_halt_comp_diversity} bins the $10$K UniProt held-out sequences into deciles based on their hydrophobic content against either Shannon entropy (left) or Jensen--Shannon (JS) divergence to uniform (right), where the JS divergence particularly remains less sensitive to rare amino acids (AAs). Both metrics tell us the same story in that they are nearly perfectly anti-correlated (Pearson $-0.99$) and both show the highest cycles to be clustered at the high-hydrophobicity end, with halt depth mostly dropping as hydrophobicity falls.

\begin{figure}[h]
    \centering
    \includegraphics[width=0.8\linewidth]{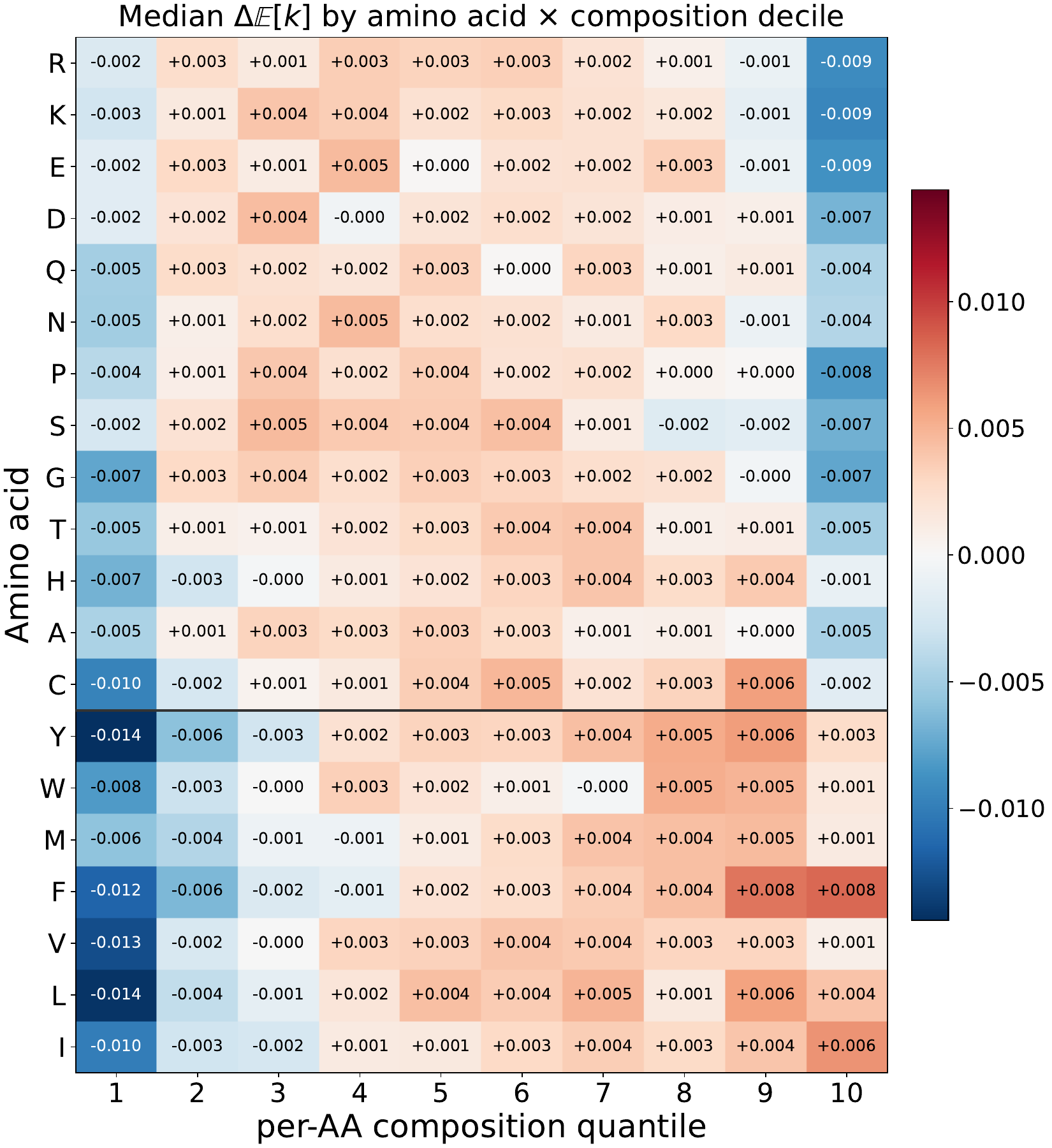}
    \caption{\textbf{Median residual halt depth $\En$ by amino acid $\times$ per-AA composition decile.} Rows are grouped into the seven hydrophobic/aromatic residues (bottom) and the remaining residues (top). The horizontal line separates the hydrophobic block (I, L, V, F, M, W, Y) from the non-hydrophobic block above. Columns are within-AA composition deciles (bin $1$: lowest fraction, bin $10$: highest). Hydrophobic AAs show increases over most of the range: heavier hydrophobic content in general drives more GMR cycles. Non-hydrophobic AAs show an inverted U-shape with extreme composition dominance driving fewer cycles.}
    \label{app:fig_halt_comp_per_aa}
\end{figure}

Figure~\ref{app:fig_halt_comp_per_aa} decomposes the effect per amino acid, with the seven hydrophobic/aromatic residues grouped at the bottom and columns being within-AA composition deciles. Each cell reports the median $\En$ residual for sequences in that AA's decile bin. The pattern splits cleanly along the horizontal separator: every one of the seven hydrophobic AAs (F, L, I, V, M, W, Y) shows an overall increase, i.e., higher composition of a hydrophobic residue drives more GMR cycles in general. By contrast, every non-hydrophobic AA (from A downward) shows an inverted U-shape, where extreme dominance of a single polar or charged residue drives \emph{fewer} recurrent cycles.

The two mechanisms are the two sides of the same question --- how many independent decisions does this residue distribution actually require? Overrepresentation of a hydrophobic residue is evolutionarily rare because random sequences with that composition would aggregate. The ones that survive do so through a specific fold or a membrane pocket, and recovering the wild-type residue at masked positions requires the model to integrate long-range structural evidence. Overrepresentation of any other residue is instead an alphabet-reduction cue, where once a polar or charged class dominates, many masked positions are predictable from local statistics and the model can commit early. The halt signal thus reads a reduced-alphabet grammar in the sense that restricted vocabularies get fewer cycles, unless the vocabulary happens to be the demanding kind.

\paragraph{Splice-isoform collection and analysis.}
\label{app:isoform_data}

For our analyses on alternative isoform sequences in Table~\ref{tab:cross_task_property_signatures}, we use curator-annotated Swiss-Prot alternative isoforms, fetched from UniProt REST (\nolinkurl{rest.uniprot.org/uniprotkb/stream}) with \texttt{includeIsoform=true} on the human accessions of our $10$K UniProt held-out split. UniProt returns for each accession the canonical sequence together with every alternative isoform Swiss-Prot curators derived from experimental transcript evidence (alternative splicing, alternative promoter usage, alternative initiation, etc., depending on the entry). We keep the isoforms as returned (no mutation or ad-hoc reconstruction). As for the held-out set, halt depth is computed on the first $512$ residues. Excluding the $50$ genes whose isoforms differ only beyond residue $512$ leaves the result unchanged ($\approx 46\%$ tighter than the size-matched null). The raw fetch returns $1{,}730$ isoform sequences across the human accessions in the held-out split. After restricting to families with at least two isoforms and applying the length- and sharpness-controlled halt-depth aggregation of \S\ref{sec:halt_biology}, we obtain $335$ genes covering $1{,}054$ measured isoforms (family-size distribution: $163$ genes with two isoforms, $83$ with three, $44$ with four, and a long right tail up to one gene with $17$). For each family we compute the within-family standard deviation of the controlled $\En$. The observed mean within-family std ($0.013$) is $46\%$ smaller than the size-matched bootstrap null over the full held-out population ($0.025$; $p<10^{-4}$ over $10^{4}$ resamples), yielding the Cohen's $d = 0.79$ reported in the main paper.

\section{Limitations}
\label{app:sec_limitation}
\begin{figure}[h]
    \centering
    \includegraphics[width=0.3\linewidth]{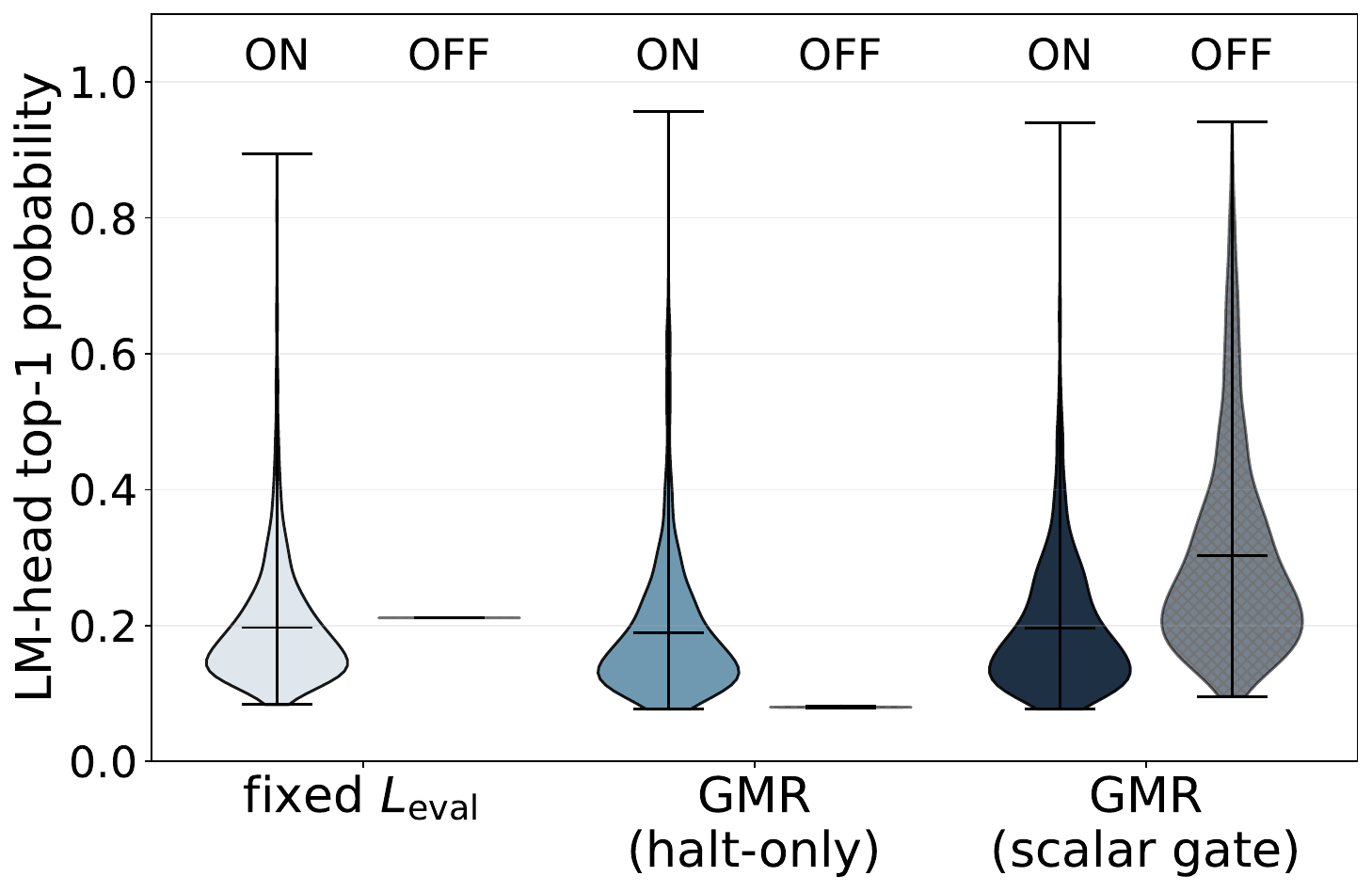}
    \caption{LM-head top-$1$ probability per real token on stability prediction, evaluated with (ON) and without (OFF) input-injection.}
    \label{fig:mech_decode_commitment}
\end{figure}
This is the first work targeting adaptive computation for test-time scaling of pLMs. As such, several caveats bound our claims. First, per-protein expected halt depth $\En$ varies within only $ \sim 1\%$ of its mean across the $10$K UniProt held-out set. GMR's adaptivity is thus a small, structured modulation around a learned mean. Second, state-level convergence is characterized empirically through vanishing update norms and residual decay. We do not establish local stability of the recurrent operator via a Jacobian analysis. Third, gates are per-sequence rather than per-token, i.e., computation adapts across proteins but not within a sequence, and moving to per-token survivals is a natural extension.
Finally, all experiments are on recAMPLIFY variants trained under the AMPLIFY data recipe, and generalization to substantially different scales, architectures, or modalities remains open.

\paragraph{The gate's stability comes at the price of reduced input sensitivity.}

Figure~\ref{fig:mech_decode_commitment} shows that removing input injection \emph{collapses} the fixed-depth and the halt-only GMR states toward a null attractor with the halt-only variant's decode particularly collapsing to a low-confidence generic token. In contrast, the scalar gate becomes more confident about a token that disagrees with its own input-injected prediction on $ \sim 90\%$ of positions. This is a specific artifact of residual gating with low degrees of freedom. With only one survival channel, the scalar gate's survival distribution collapses to an input-invariant trajectory. The gate thus buys representational stability by substituting a stored default in place of a truly input-driven recurrence.

\end{document}